\PassOptionsToPackage{table}{xcolor}
\documentclass[11pt]{article}
\usepackage[letterpaper,margin=1in]{geometry}
\usepackage{iftex}
\usepackage{arena-brand}
\usepackage{natbib}
\setcitestyle{authoryear,round,citesep={;},aysep={,},yysep={;}}

\usepackage{microtype}
\usepackage{url}
\usepackage{booktabs}
\usepackage{graphicx}
\usepackage{amsmath,amssymb}
\usepackage{array}
\usepackage{tabularx}
\usepackage{threeparttable}
\usepackage{makecell}
\usepackage[table]{xcolor}
\usepackage{multirow}
\usepackage{fontawesome5}
\usepackage{hyperref}
\usepackage{wrapfig}
\usepackage[breakable,skins,listings]{tcolorbox}
\usepackage{titlesec}
\titleformat{\section}{\ArenaHeadingFont\color{ArenaInk}\fontsize{16}{19}\selectfont}{\thesection.}{0.6em}{}
\titleformat{\subsection}{\ArenaHeadingFont\color{ArenaInk}\fontsize{13.5}{16}\selectfont}{\thesubsection.}{0.6em}{}
\titleformat{\subsubsection}{\ArenaHeadingFont\color{ArenaInk}\fontsize{12}{14.5}\selectfont\bfseries}{\thesubsubsection.}{0.6em}{}
\titlespacing*{\section}{0pt}{2.6ex plus 0.5ex minus 0.2ex}{1.0ex}
\titlespacing*{\subsection}{0pt}{2.0ex plus 0.4ex minus 0.2ex}{0.7ex}
\titlespacing*{\paragraph}{0pt}{1.25ex}{1em}
\newcommand{\wccomment}[1]{\textcolor{orange}{}}
\newcommand{\chocomment}[1]{\textcolor{orange}{}}
\newcommand{\skcomment}[1]{\textcolor{olive}{}}
\newcommand{\cdcomment}[1]{\textcolor{magenta}{}}
\newcommand{\todo}[1]{\textcolor{red}{}}

\newcommand{\model}[1]{\texttt{#1}}

\newcommand{\method}{Interaction-as-a-Judge}
\newcommand{\preferences}{ArenaWebdevPreferences}
\newcommand{\prefsize}{$6,858$}
\newcommand{\prefreleasesize}{$6,390$}
\newcommand{\correctness}{ArenaWebdevCorrectness}
\newcommand{\correctsize}{$222$}
\newcommand{\correctreleasesize}{$189$}
\newcommand{\requirementsize}{$412$}
\newcommand{\turnsize}{$828$}
\newcommand{\tracesize}{$67$}
\newcommand{\HFLogo}{\raisebox{-0.12em}{\includegraphics[height=0.95em]{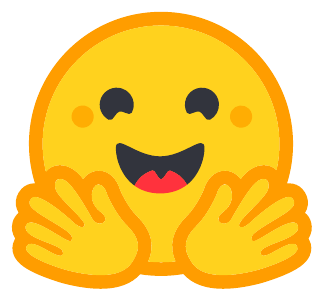}}}
\newcommand{\arena}{Arena}
\newcommand{\webdev}{Code Arena}
\newcommand{\agentmode}{Agent Mode}

\definecolor{DiagGray}{gray}{0.42}

\title{Interaction-as-a-Judge}

\author{%
\mbox{Wayne Chi\textsuperscript{1,2}\thanks{%
  \protect\begin{tabular}[t]{@{}l@{}}
    Work done as part of an internship at Arena.\\
    Correspondence: 
    \texttt{waynechi@andrew.cmu.edu;}
    \texttt{\{ihung,chojui\}@arena.ai}
  \end{tabular}%
}},
\mbox{I-Hung Hsu\textsuperscript{2}},
\mbox{Li Chen\textsuperscript{2}},
\mbox{Wei-Lin Chiang\textsuperscript{2}},
\mbox{Ion Stoica\textsuperscript{2,3}},
\mbox{Anastasios Nikolas Angelopoulos\textsuperscript{2}},
\mbox{Ameet Talwalkar\textsuperscript{1}},
\mbox{Chris Donahue\textsuperscript{1}},
\mbox{Cho-Jui Hsieh\textsuperscript{2,4}}
}

\newcommand{\affiliations}{%
\mbox{\textsuperscript{1}\,\href{https://cmu.edu}{Carnegie Mellon University}}\quad
\mbox{\textsuperscript{2}\,\href{https://arena.ai}{Arena}}\quad
\mbox{\textsuperscript{3}\,\href{https://berkeley.edu}{University of California, Berkeley}}\quad
\mbox{\textsuperscript{4}\,\href{https://ucla.edu}{University of California, Los Angeles}}
}

\makeatletter
\renewcommand{\@maketitle}{%
  {\raggedright\setlength{\parindent}{0pt}%
    \vspace*{-0.5in}%
    \hbox to \linewidth{\ArenaLogo[0.256in]\hfil
      \raisebox{3pt}{\ttfamily\fontsize{9}{11}\selectfont\color{ArenaTextSecondary}October 23, 2026}}\par
    \vspace{8pt}
    {\color{ArenaInk}\rule{\linewidth}{1pt}}\par
    \vspace{16pt}
    {\ArenaTitleFont\fontsize{28}{33}\selectfont\color{ArenaInk}\@title\par}
    \vspace{12pt}
    {\fontsize{11.5}{14}\selectfont
      \def\@makefnmark{\hbox{\@textsuperscript{\normalfont\@thefnmark}}}%
      \@author\par}
    \vspace{6pt}
    {\fontsize{9.5}{11.5}\selectfont\hypersetup{urlcolor=ArenaInk}\affiliations\par}%
  }
}
\makeatother
\renewenvironment{abstract}{%
  {\noindent\ArenaHeadingFont\large\bfseries Abstract\par}
  \vspace{8pt}\normalfont\normalsize
  \setlength{\parindent}{0pt}%
}{\par}

\begin{document}

\suppressfloats[t] 

\maketitle
\begin{tcolorbox}[enhanced, colback=ArenaPaper, colframe=ArenaBorderFaint,
  boxrule=0.75pt, arc=6pt, boxsep=0pt,
  left=12pt, right=12pt, top=12pt, bottom=12pt,
  before skip=16pt, after skip=12pt]
\begin{abstract}
Coding agents are tasked with producing sophisticated interactive applications such as websites, desktop applications, and games. While humans assess these applications interactively, current automated LLM evaluations primarily focus on code analysis and unit tests. This mismatch can cause LLM judges to miss failures that are immediately apparent to human users. We study \method{}, an evaluation paradigm where computer-use agents (CUAs) directly interact with an application. We demonstrate that \method{} can improve both \emph{evaluation} and \emph{generation} of interactive applications. To this end, we collect two complementary datasets: $6,858$ pairs of applications with human preferences and corresponding agent traces, and $222$ applications with human-verified requirement labels. We publicly release $6,390$ and $189$ of these, respectively. Relative to standard non-interactive approaches, we find that \method{} improves evaluation in terms of predicting human preferences ($65.6\% \to 69.1\%$) and requirement verification ($77.0\% \to 81.6\%$). We also observe scaling benefits: across both preference alignment and requirement verification, higher-quality interactions improve performance. Finally, through both inference- and training-time experiments, we demonstrate that \method{} can also improve the quality of generated artifacts. At inference time, \method{} repairs $75.5\%$ of broken requirements within five attempts, compared to $55.4\%$ for code-only self-feedback. At training time, using \method{} as a reward signal improves win-rate over the base model by $6.7$pp while reducing both generated code length by $35.7\%$ and build failure rates by $7.2$pp compared to a code-only reward signal. 

\end{abstract}
\vspace{10pt}
{\noindent
\begin{tabular}{@{}c@{\hspace{0.65em}}l@{\hspace{1.25em}}l@{}}
  \faGlobe & \textbf{Website} & \url{https://arena.ai/blog/interaction-as-a-judge}\\
  \HFLogo & \textbf{Datasets} & \href{https://huggingface.co/datasets/lmarena-ai/arena-webdev-preferences}{\nolinkurl{preferences}}\enspace{\color{ArenaTextSecondary}\textperiodcentered}\enspace\href{https://huggingface.co/datasets/lmarena-ai/arena-webdev-correctness}{\nolinkurl{correctness}}
\end{tabular}\par}
\end{tcolorbox}
\begingroup
\renewcommand{\thefootnote}{}%
\footnotetext{%
  \begin{tabular}[t]{@{}l@{}}
    \url{https://huggingface.co/datasets/lmarena-ai/arena-webdev-preferences}\\
    \url{https://huggingface.co/datasets/lmarena-ai/arena-webdev-correctness}
  \end{tabular}%
}%
\endgroup

\section{Introduction}
\label{sec:intro}

As coding agents improve, so does the complexity of the software artifacts they produce and, in turn, the difficulty of evaluating those artifacts.
Correctness and preference are often difficult to assess from code alone.
For example, a game may compile while still failing behaviorally: player animations may not match movement, scene transitions may freeze the application, or interactions may lack the visual and temporal feedback needed to feel responsive and intentional.
A natural method for evaluation---and the way most humans evaluate software---is simply to interact with and visually inspect the final product.
Yet most current LLM-as-a-Judge~\citep{zheng2023judging, chiang2024chatbot} and Agent-as-a-Judge~\citep{zhuge2024agentasajudge, webdevjudge2025} methods still rely on static or unimodal inspection of software artifacts.
Properly evaluating such artifacts requires moving beyond static inspection to active interaction.

\begin{figure}[t]
    \centering
    \includegraphics[width=\linewidth]{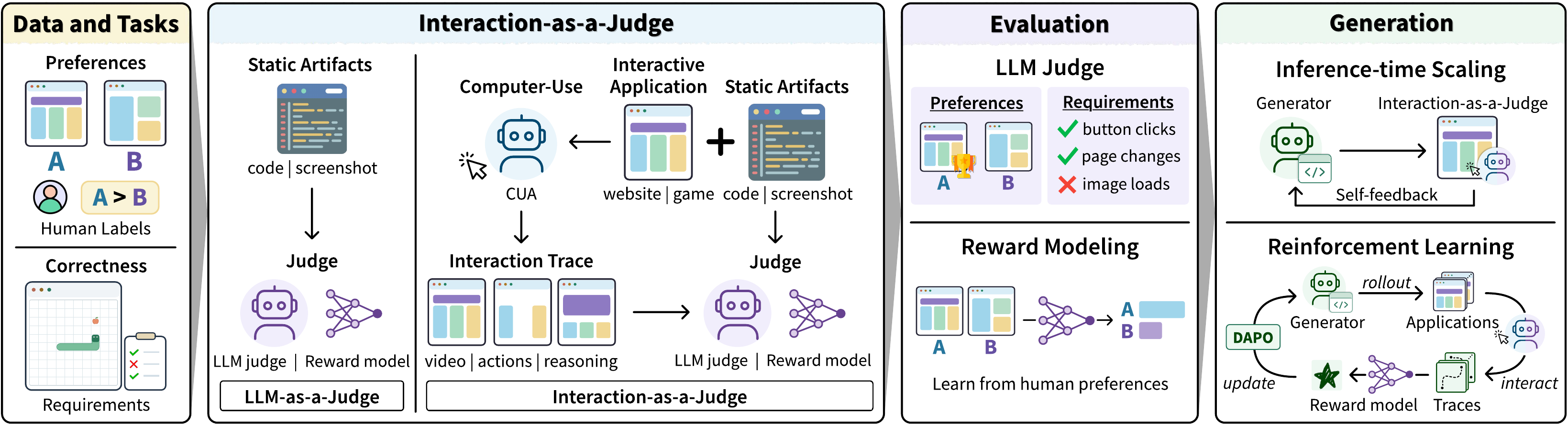}
    \caption{We present \method{}, an evaluation paradigm that uses interaction traces to improve evaluation and generation.
    We compare against non-interactive judges and evaluate on pairwise preferences and reward modeling.
    We also show how to apply \method{} for inference-time scaling and as a reward signal for standard post-training via reinforcement learning.}    
    \label{fig:figure1}
\end{figure}

In this work, we study the use of interaction traces to improve evaluation accuracy and in turn improve generation quality.
Concretely, a computer-use agent (CUA) interacts with an application and produces a multimodal interaction trace which can be used to evaluate the quality of the application, or optionally, provide feedback for improving it.
We call this evaluation paradigm \textit{``\method{}''}, by analogy with LLM-as-a-Judge~\citep{zheng2023judging}.
While some prior work has explored this paradigm in different application settings~\citep{bian2025click,lu2025webgenagentenhancinginteractivewebsite,wang2026verificationhorizonsilverbullet},
we are the first to systematically study it in a controlled fashion. 
Specifically, we investigate how dimensions such as interaction strategy and agent capability affect evaluation accuracy for both pairwise human preference prediction and correctness verification. 
We further show that these evaluation signals can improve generation using simple and standard methods at both inference and training time, namely inference-time scaling and reinforcement learning with a learned reward model.

To study \method{}, we introduce two datasets: \preferences{} and \correctness{}.
\preferences{} is a set of \prefsize{} agentic code traces across seven different categories of web development (games, simulations, reference-based designs, content-creation tools, brand sites, consumer apps, and data applications).
Alongside the agent traces, each data point contains a prompt, two websites generated from the prompt, and labeled human preferences.
\correctness{} is a set of \tracesize{} high-quality game development traces with \turnsize{} conversation turns.
We hand-label \requirementsize{} requirements across various versions of each game as the game develops by interpreting requirements from a user's prompt and feedback.
Requirements are then hand-verified as  \textit{working} or \textit{broken} depending on if the requirement is successfully implemented.

We find that \method{} increases alignment with human preferences for both in-context LLM judges and trained reward models.
On \preferences{}, adding interaction traces on top of code significantly improves overall in-context LLM judge accuracy by up to $3.5$ percentage points ($65.6\%$ to $69.1\%$) and by $1.5$ percentage points ($67.6\%$ to $69.1\%$) when compared to adding only static visual information.
Fine-tuning a reward model with interaction traces improves accuracy from $68.2\%$ to $70.2\%$ and outperforms in-context LLM judges.
We find that higher \textit{quality} computer-use results in the best overall performance, although even a random heuristic can capture non-trivial gains.
On \correctness{}, \method{} improves requirement verification accuracy from $77.0\%$ for a code-only Agent-as-a-Judge~\citep{zhuge2024agentasajudge} to $81.6\%$. 
\method{} also scales with model capability: accuracy increases from $76.4\%$ with GPT 5.6 Luna to $81.6\%$ with GPT 5.6 Sol when each serves as both the CUA and the judge. 

We show that these evaluation gains also improve generation.
At inference-time, we use \method{} in a self-feedback loop where an agent iteratively attempts to repair broken requirements and evaluates its repairs using \method{}.
Compared with code-only feedback, adding interaction to the feedback mechanism increases the percentage of initially broken requirements successfully repaired within five attempts from $55.4\%$ to $75.5\%$.
We also use a reward model trained solely on interaction traces, without code, to post-train a generation model through reinforcement learning. Relative to using a code-only reward model, this method reduces the build failure rate by $7.2$ percentage points and reduces code-context growth---a known problem in code reward models~\citep{wang2026verificationhorizonsilverbullet}---by $35.7\%$, while achieving comparable generation quality.
Additionally, we show that websites generated by this post-training recipe attain a $6.7$pp higher win rate against the base model.

In summary, our contributions are as follows:

\begin{itemize}
    \item We study \method{} for both preference judgements and correctness verification and release two datasets---\preferences{} and \correctness{}---for further research into both \method{} and agentic generation of interactive applications more broadly.
    \item We show that \method{} improves evaluation capabilities.
    \method{} improves overall alignment with human preferences for both in-context LLM judges and trained reward models.
    Additionally, \method{} improves an agent's ability to verify requirements when compared to a code-only agentic judge.
    \item We demonstrate how \method{} can be used to improve generation through both inference-time scaling and reinforcement learning training.
    \method{} improves the number of successful repairs in an iterative self-feedback loop and improves trained model quality when used as a reward signal.
\end{itemize}

\section{Tasks and Datasets}

We study \method{} under two complementary settings: human preferences and correctness verification.
Human preferences are useful when quality is subjective or difficult to capture with explicit requirements, while correctness verification is appropriate when desired behavior can be specified objectively.
In many applications, both are necessary.
For example, a software application may satisfy its functional requirements while still differing substantially in usability or visual quality.
The inverse can also be true, where an application is visually preferable, but functionally broken.
Effective evaluation methods should therefore be able to assess both correctness and preference.


\subsection{Preferences}
For pairwise preferences, a judge is given a shared prompt and two candidate outputs and must predict which output would be preferred by a human user~\citep{zheng2023judging}.
We focus on pairwise comparisons in this work rather than other preference tasks such as single answer grading (e.g., predicting a Likert scale score),
but our approach is not restricted to pairwise comparisons.

\paragraph{\preferences{}.} 
We release \preferences{}, a dataset of \prefsize{} pairwise comparisons across seven categories of web development.
We collect our dataset from \arena{}'s \webdev{} where real users prompt two hidden models and vote for their preferred website.
Each datapoint in \preferences{} contains the user's prompt, both agentic traces, both final websites, and the ground truth user label.
We categorize websites into games, simulations, reference-based designs, content-creation tools, brand sites, consumer product, and data applications following the original leaderboard categories.
To ensure that the dataset is high quality, we filter such that all ties are removed, websites are built with React, outputs build successfully, outputs contain $\geq 5\%$ character difference, and users spent at least 60 seconds and up to 10 minutes interacting with the websites.
All personally identifiable information (PII) has been de-identified through a PII filtering process.
We evaluate on all \prefsize{} comparisons and, due to internal filtering policies, publicly release \prefreleasesize{} of them.
To our understanding, this is the largest collection of pairwise agentic web development traces and websites released for open source use.
We provide further dataset details in Appendix~\ref{app:data}.

\subsection{Correctness}
We consider two types of correctness verification: \emph{requirement verification} and \emph{requirement discovery}.
In requirement verification, the judge is given an application, its code, and a set of explicit requirements, such as \textit{``Mario must defeat a Goomba when jumping on it.}''
The judge must determine whether each requirement is working or broken for that application.
In requirement discovery, the judge is given the application and its code but no explicit list of requirements.
It must first identify the relevant requirements automatically and then determine whether those requirements are working.

\paragraph{\correctness{}.}
We create \correctness{}, a dataset containing \tracesize{} game development traces with \turnsize{} user turns and labeled with \requirementsize{} requirements across \correctsize{} versions of various games.
We collect our dataset from \arena{}'s \agentmode{} where users can request a random agent to do tasks such as web search, software development, or image generation.
We focus on game development as games are highly interactive and also challenging software development tasks~\citep{chi2026gamedevbenchevaluatingagenticcapabilities}.

To generate requirements from user traces, we analyze the user trace for specific complaints, and extract the complaint into a broken requirement which is linked to a specific version of the game matching when the user complaint occurred.
For example, a user complaint indicating that the player falls through platforms would be annotated as a requirement indicating that the player must be able to stand on platforms.
We use GPT 5.6 Luna to label, analyze, and then annotate each complaint as a requirement.
A human annotator then reviews each game to verify that the requirements are sensible and broken in the game version.
To extract working requirement labels, we take our existing broken requirements, and check to see if the user either explicitly confirms a fix or does not revisit this complaint.
We use GPT 5.6 Luna for initial annotation and verify via human annotation.
In total, we annotate 323 broken and 89 working requirements.
We evaluate on all \correctsize{} game versions and publicly release \correctreleasesize{} of them.
We provide further dataset details in Appendix~\ref{app:data}.

\section{Interaction-as-a-Judge}
\label{sec:method}

\begin{figure}
    \centering
    \includegraphics[width=0.9\linewidth]{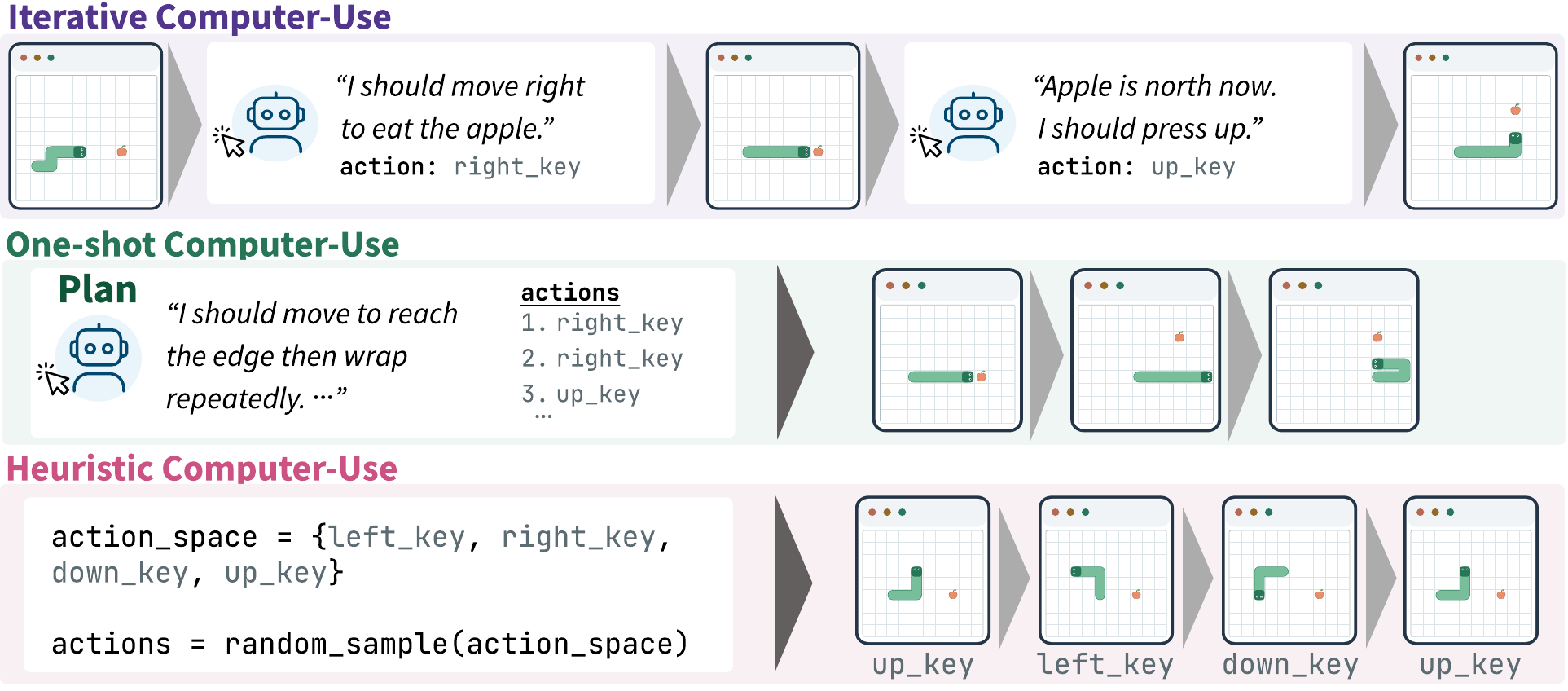}
    \caption{We study three approaches to computer-use. Iterative computer-use is analogous to standard frame-by-frame computer-use. One-shot computer-use plans once at the start to save costs. Heuristic computer-use uses a pre-determined algorithm (e.g., random) to choose actions.}
    \label{fig:method}
\end{figure}

\method{} is an evaluation paradigm which uses computer-use interaction traces to improve downstream evaluation accuracy.
\method{} consists of an \textit{interaction} component and a \textit{judge} component.
These two components can be either the same LLM, different LLMs, or a mixture of LLMs and other methods. 
For example, a cheap computer-use heuristic paired with an LLM judge, or an LLM-based CUA paired with a trained reward model.

\subsection{Interaction}
Following~\citet{koh2024visualwebarena}, we
model interaction with the environment as a partially observable Markov decision process (POMDP).
We define the environment as $\mathcal{E} = (\mathcal{S}, \mathcal{A}, \Omega, f)$ where $\mathcal{S}$ is the state space, $\mathcal{A}$ is the action space, and $\Omega$ is the observation space.
In our work, $o_t \in \Omega$ is a visual representation of the application (a screenshot), but the Document Object Model (DOM) is also commonly used.
An action $a_t \in \mathcal{A}$ induces a deterministic transition $f  : \mathcal{S} \times \mathcal{A} \rightarrow \mathcal{S}$.
To interact with the environment, we define an interaction policy $\pi$ that determines the sequence of actions taken in the environment. 
Executing $\pi$ in the environment produces an interaction trace $\tau = ((o_t,\rho_t,a_t))_{t=0}^{T}$, where $o_t$, $\rho_t$, and $a_t$ denote the observation, agent reasoning, and action at timestep $t$, respectively. We allow $\rho_t = \varnothing$ when no explicit reasoning is produced.
Actions include standard computer inputs such as moving the mouse, clicking, and typing---the full action space is in Appendix~\ref{app:action-space}.

There are multiple approaches to interact with a software application environment, with the primary trade-off being between adaptivity and inference cost. We consider three interaction strategies that vary in how frequently they query a model and observe the resulting environment state.

\paragraph{Iterative computer-use.}
The most general approach is to use a computer-use agent that iteratively observes and acts in the environment. 
We represent the agent as a policy $\pi_\theta^{\mathrm{iter}}$, where at timestep $t$ it selects an action $(a_t,\rho_t) \sim \pi_\theta^{\mathrm{iter}}(\tau_{<t}, i)$, where $\tau_{<t}$ is the interaction trace prior to timestep $t$, $\rho_t$ is the reasoning for action $a_t$, and $i$ are instructions to the computer-use agent.
For preferences this might be a broad instruction to investigate the application to determine quality while for correctness the goal could be a specific requirement it must verify.
The environment then transitions to $s_{t+1} = f(s_t, a_t)$, producing a new observation that can inform the next action. 
This iterative formulation allows the agent to adapt as the environment changes and corresponds to the standard computer-use setting. 
However, it can be prohibitively expensive, as it requires a model query at every interaction step. 

\paragraph{One-shot computer-use.}
A sufficiently capable agent could in principle predict the sequence of actions to take from only the initial state without observing the intermediate outcomes of its actions.
We represent this as a policy $\pi_\theta^{\mathrm{one}}$ that, given the initial observation $o_0$, reasons and then predicts an entire action sequence $(a_{0:T}, \rho) = \pi_\theta^{\mathrm{one}}(o_0, i)$ to be executed sequentially in the environment up to timestep $T$.
This requires only a single model query, substantially reducing inference cost~\citep{wang2026verificationhorizonsilverbullet}, but prevents the agent from adapting when the environment evolves differently from its prediction.
When application code $c$ is available, we additionally provide it to the agent for planning, giving $(a_{0:T}, \rho) = \pi_\theta^{\mathrm{one}}(o_0,i, c)$.

\paragraph{Heuristic computer-use.}
The cheapest alternative is to forgo model-based action selection entirely.
In this work, we consider a simple heuristic policy $\pi^{\mathrm{heur}}$ that selects an action uniformly at random.
To make this interaction more effective, we restrict the policy to a candidate set $\mathcal{C}(o_t) \subseteq \mathcal{A}$ of interactions that are likely to induce meaningful changes in the environment, such as clicking buttons or typing in text boxes, giving $a_t \sim \pi^{\mathrm{heur}}(o_t) = \mathrm{Unif}(\mathcal{C}(o_t))$.
Because this incurs no model inference cost, the policy can select an action after every environment transition.
Despite its simplicity, this can produce substantial interaction when $\mathcal{C}(o_t)$ can be effectively identified, but is less effective for environments such as games, where meaningful interactions are difficult to enumerate automatically.

\subsection{Judging}

After the interaction trace $\tau$ is generated, the judge uses the trace to inform its downstream judgement.
We study both an off-the-shelf LLM, which we use for both predicting human preferences and verifying correctness, and a trained reward model, which we use for predicting human preferences.
A key benefit of using off-the-shelf LLMs is that it allows interaction-as-a-judge to leverage 
frontier capabilities without needing to train a custom model,  which is often infeasible without data and compute resources.
However, a reward model trained on sufficient in-domain data can achieve higher accuracy and reduce inference costs.

\paragraph{LLM Judge.}
For human preferences, an off-the-shelf LLM judge takes in the user request, source code for each website, and an interaction trace for each website and outputs a judgement indicating which website is better.
Due to known position bias~\citep{zheng2023judging}, every battle is judged twice with sides swapped (AB and BA) and accuracy is the mean judge-human agreement of both passes.
For requirement verification, the judge takes in the user request, the source code, the interaction trace, and a requirements list and labels each requirement as working, broken, or unverifiable.
Requirement discovery is similar except no requirement list is given and the judge must output a list of potential requirements itself.
For judges that accept video natively, we convert $o_{0:T} \in \tau$ to video; otherwise, we pass in each frame as an image.
Actions and reasoning are passed in directly as text sequences.

\paragraph{Reward Model.}
We train our reward model $r_\theta$ by adding a scalar reward head and optimizing over a standard Bradley--Terry loss~\citep{bradley1952rank}:


$$\mathcal{L}_{\mathrm{BT}}(\theta) = -\mathbb{E}_{(x, y_w, y_l, \tau_w, \tau_l) \sim \mathcal{D}} \left[ \log \sigma\!\left( r_\theta(x, y_w, \tau_w) - r_\theta(x, y_l, \tau_l) \right) \right]$$

where $x$ is the user prompt, $y_w$ and $y_l$ are the source code for the winner and loser respectively, and $\tau_w$ and $\tau_l$ are the traces.
This is similar to standard reward model training~\citep{christiano2017deep,
ouyang2022training} with the addition of interaction trace information.

%

\section{
\method{} for Evaluation
}
\label{sec:evaluation-results}


We measure the extent to which evaluation via \method{} aligns with human preferences and verified correctness.
In our main experiments, all interaction traces consist of 30 actions.
For iterative and one-shot computer-use, we use GPT 5.6 Luna with medium reasoning as our computer-use agent.
Unless specified, \method{} includes code as input.
Games differ from other categories due to their real-time input and difficulty in identifying meaningful candidate interactions (e.g., lack of buttons).
We describe our harness, action space, and changes for games in Appendix~\ref{app:harness}.

\subsection{Human Preferences}
\label{sec:human-preferences}

\begin{figure*}[t]
    \centering
    \includegraphics[width=\textwidth]{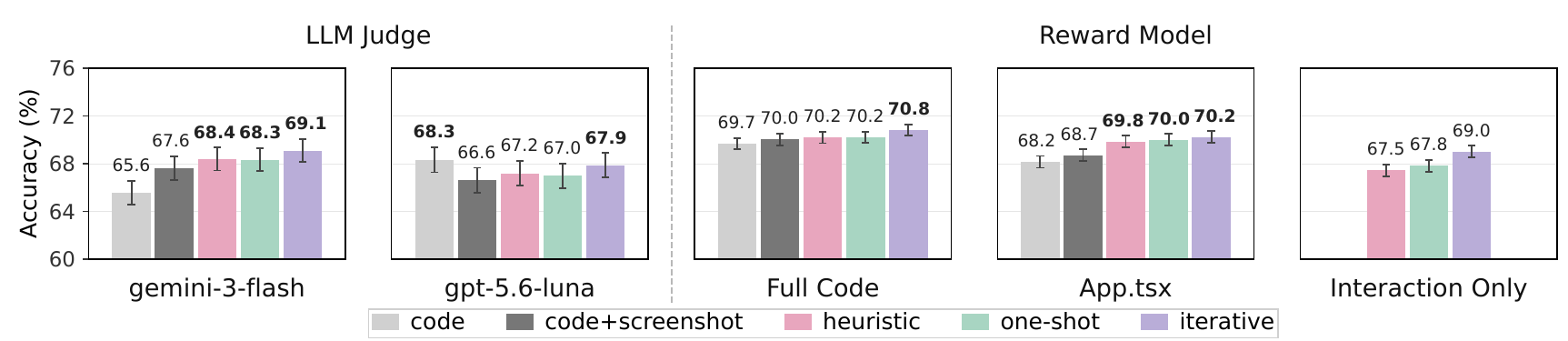}
    \caption{We train a reward model with varying levels of interaction and observe that including interaction data improves accuracy in terms of alignment with human preferences.
    GPT 5.6 Luna is the only exception where adding interaction offers comparable performance to code only.
    We fine-tune from Qwen-3.5-9B as our base model and train each approach 5 times.
    Bolded values are statistically significant against code+screenshot, calculated by a sample-wise paired t-test.
    \wccomment{Add Qwen-3.5-27B}
    }
    \label{fig:main}
\end{figure*}

We evaluate the effectiveness of \method{} for judging human preferences using \preferences{} (Figure~\ref{fig:main}). 
We compare \method{} against two static-evidence judges: a code-only judge and a code and screenshot judge.
The code-only judge receives only the code as input, making it functionally identical to either LLM-as-a-Judge~\citep{zheng2023judging} or a standard code reward model.
The code and screenshot judge receives the code and a full page screenshot of the landing page. 
This same screenshot is also given as input to \method{} approaches. 
We also vary the amount of code received between the full code capped at $32$k tokens and just the core App.tsx file to study the effects of code on \method{}.
Additional information such as category breakdowns can be found in Appendix~\ref{app:data}.

To train our reward model, we fine-tune Qwen-3.5-9B~\citep{qwenteam2026qwen35omnitechnicalreport} as our base model and train on a private dataset of $20,160$ human preference pairs.
This dataset is structurally identical to \preferences{}, but is temporally disjoint consisting of data prior to \preferences{}, mirroring how data collection occurs in real settings.
This means that the reward model is evaluated on newer, more advanced outputs not seen in the training data.
For the visual component of the interaction trace, we input it directly as video and freeze the video encoder weights during training.

For our LLM judge, we use Gemini 3 Flash and GPT 5.6 Luna with xhigh reasoning.
Both were chosen due to their cost-effectiveness with Gemini 3 Flash being chosen due to its native video capabilities and GPT 5.6 Luna being the main computer-use model used throughout the paper.
To account for ordering bias in LLM-as-a-Judge approaches, we evaluate both orderings and take the average of the results.

\textbf{\method{} increases alignment with human preferences.}
We find that for both a trained reward model and an LLM Judge, adding interaction improves judge alignment with human preferences (Figure~\ref{fig:main}).
Adding visual information through a screenshot improves performance over code and adding interaction improves performance further.
We see the largest gap in Gemini 3 Flash as a judge where accuracy improves from $65.6\% \rightarrow 69.1\%$ between code only and iterative \method{}.
GPT 5.6 Luna as a judge is the sole exception where we do not see any statistical differences between any approach.
We speculate that for pairwise preferences, GPT 5.6 Luna is both optimized for code and also unable to properly utilize visual information given its strong code-only performance and degradation when adding even a single screenshot.
Finally, our trained reward model ($70.8\%$) is able to outperform an off-the-shelf LLM judge ($69.1\%$).

\textbf{Interaction alone is sufficient information for training a reward model.}
We find that training a reward model on the interaction traces alone (without code) provides comparable performance to a code reward model.
Iterative computer-use without code performs only slightly worse than a code reward model trained on the entire code context ($69.0\%\  \text{vs}\ 69.7\%$), and performs comparably to \method{} with code when using Gemini 3 Flash as a judge.
This opens doors to training reward models on data where the application is accessible, but code is inaccessible. 
In general, \method{} judgements remain robust as code context decreases, seeing small decreases between seeing the full source code versus only the core App.tsx file.
However, both one-shot and heuristic computer-use degrade more compared to iterative computer-use when no code is available.

\textbf{Interaction quality affects alignment with human preferences.}
We observe that iterative computer-use consistently performs the best across all experiments.
Due to its ability to respond to new state changes in the environment, iterative computer-use generates the highest-quality traces among the three approaches.
Additionally, we find that higher-quality interaction at inference can compensate for lower-quality traces at training time.
We take our reward model trained on heuristic computer-use and run inference using iterative computer-use instead.
We find that performance increases across almost every single category (Appendix~\ref{app:cross-train}, Figure~\ref{fig:cross-train}).
The inverse is not true---a reward model trained on iterative computer-use degrades when using heuristic computer-use traces at inference time.

%


\subsection{Correctness Verification}

\begin{figure}[t]
    \centering
    \includegraphics[width=0.9\textwidth]{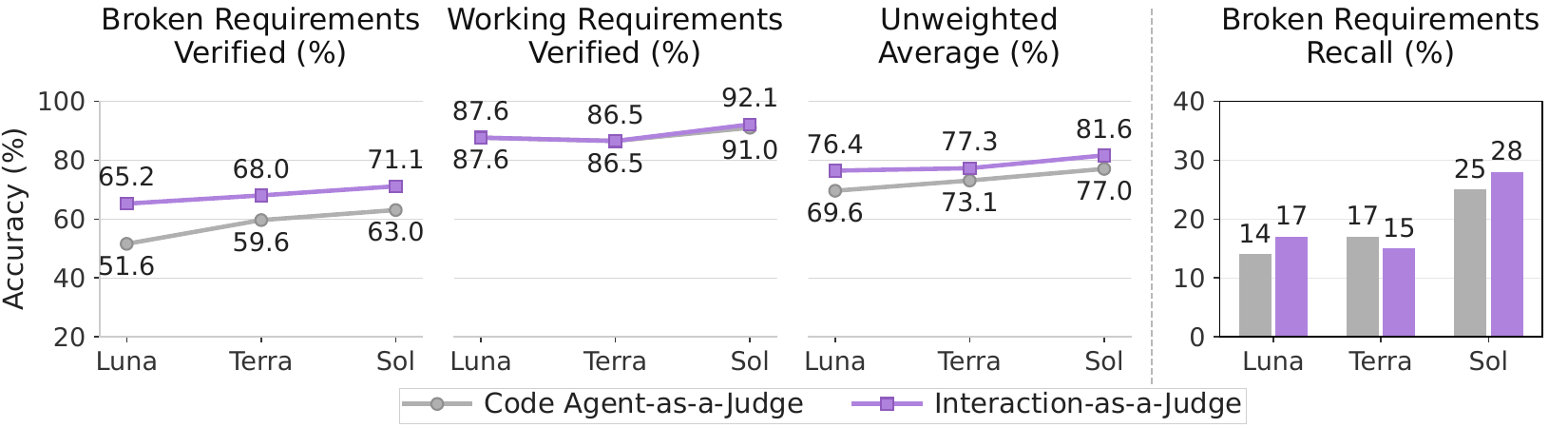}
    \caption{We evaluate \method{}'s ability for requirement verification (left) and requirement discovery (right).
    For requirement verification, we see that \method{} is better than a code only Agent-as-a-Judge, especially at identifying broken requirements. 
    Additionally, we observe that performance scales as model capabilities improve.
    We find that requirement discovery is still challenging for both approaches.
    }
    \label{fig:correctness}
\end{figure}

We evaluate \method{} on its ability to both verify and discover requirements using our \correctness{} dataset (Figure~\ref{fig:correctness}).
For requirement verification, we give the agent the code, the game as an interactable application, and the list of requirements which can be either all broken or all working.
The agent is asked to label each requirement as working, broken, or unverifiable.
For requirement discovery, we focus only on broken requirements and ask the agent to list out all possible broken requirements that require repair without passing in the original requirements list. 
We compare \method{} against a code only Agent-as-a-Judge~\citep{zhuge2024agentasajudge}.
We use the GPT 5.6 model family (Luna, Terra, Sol) all with medium reasoning to measure varying model capabilities.

\textbf{\method{} outperforms code only requirement verification.}
We find that across all models, \method{} performs better than a code only Agent-as-a-Judge.
This is primarily because \method{} is significantly better at verifying broken requirements ($71.1\%$) compared to a code only Agent-as-a-Judge ($63.0\%)$.
We find that both approaches perform comparably for verifying working requirements at $92.1\%$ and $91.0\%$ for \method{} and code-only Agent-as-a-Judge respectively.

\textbf{\method{} scales with model capability.}
We see in Figure~\ref{fig:correctness} that broken requirement verification performance continues to improve as the overall capability of the model improves going from $65.2\%$ with GPT 5.6 Luna to $68.0\%$ with GPT 5.6 Terra and $71.1\%$ with GPT 5.6 Sol.


\section{
\method{} for Generation
}

\subsection{Inference-time Scaling}

\begin{figure}[h!]
    \centering
    \includegraphics[width=\linewidth]{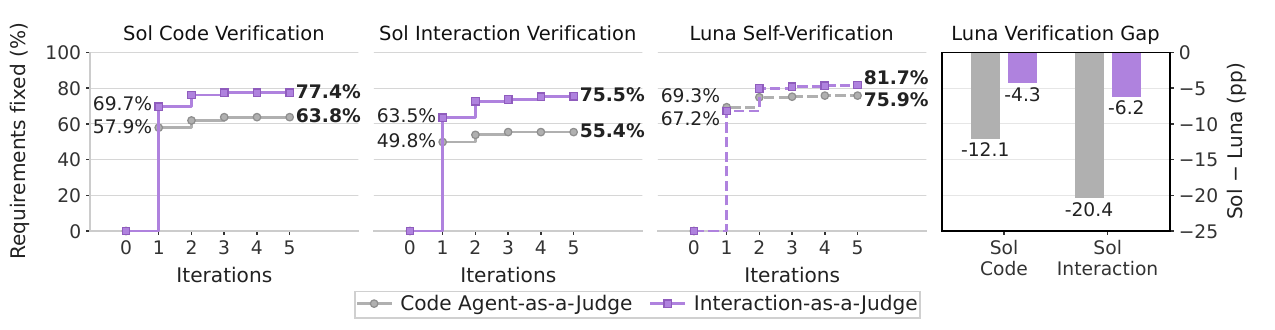}
    \caption{We compare \method{} against a code-only Agent-as-a-Judge at iterative self-feedback and requirement repair.
    \method{} repairs $75.5\%$ of initially broken requirements compared to $55.4\%$ repaired by the code-only Agent-as-a-Judge. 
    We see this improvement even when evaluating using Sol as a code-only Agent-as-a-Judge.
    Additionally, the gap between self-reported accuracy and accuracy measured by GPT 5.6 Sol is less for \method{}.}
    \label{fig:inference}
\end{figure}

We evaluate whether interaction-based self-feedback improves an agent's ability to repair broken requirements (Figure~\ref{fig:inference}).
An effective inference time strategy is to iteratively repair and verify the resulting application.
Regardless of implementation capabilities, self-verification capability becomes important for determining whether the repairs are correct or incorrect.
We use the games with broken requirements in \correctness{} as our task list.
We ask the agent to repair the broken requirements for each game and self-verify across five iterations.
We compare \method{} against a code only Agent-as-a-Judge and use GPT 5.6 Luna with medium reasoning as the agent for both approaches.
After all iterations are complete, we use a more capable model (GPT 5.6 Sol) with \method{} and code-only Agent-as-a-Judge to determine if the repairs were successful or not.
We measure success per requirement and report repair success rate after each iteration.

\paragraph{\method{} repairs more broken requirements than a code only Agent-as-a-Judge.}
After five iterations, \method{} successfully repairs $75.5\%$ of initially broken requirements, compared to $55.4\%$ for code-only self-feedback, an improvement of $20.1$ percentage points (Figure~\ref{fig:inference}, middle).
This improvement persists even when using GPT 5.6 Sol as a code-only Agent-as-a-Judge (Figure~\ref{fig:inference}, left), highlighting the improvements \method{} provides.
We see in Figure~\ref{fig:inference} that \method{} not only improves faster but also outperforms the code-only judge within its first iteration.
Additionally, the self-reported accuracy gap between GPT 5.6 Luna and GPT 5.6 Sol is significantly less for \method{} compared to a code only Agent-as-a-Judge (Figure~\ref{fig:inference}, right), suggesting a lower false positive rate when using \method{}.

\subsection{Training}

\begin{figure}[t!]
    \centering
    \includegraphics[width=\linewidth]{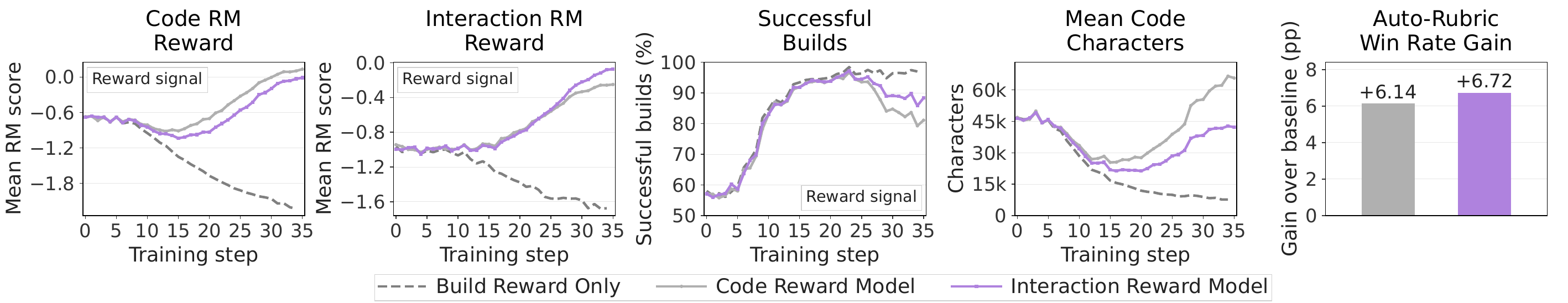}
    \caption{We train a Qwen-3.5-9B model using 
    code-only and interaction-only 
    reward models. 
    As expected, each performs better on its respective training objective, but the interaction-only model leads to improvements in neutral criteria such as build success rate and decreased code length. 
    We use an auto-rubrics based judge and find a significant improvement over the base model.
    The build reward baseline naturally degrades in reward for both reward models.
    }
    \label{fig:training}
\end{figure}

We study the potential 
of \method{} at training time to improve downstream performance in generating interactive applications. 
Specifically, 
we explore using an 
interaction-only reward model as the reward signal for post-training via reinforcement learning (Figure~\ref{fig:training}).
We start our training from Qwen-3.5-9B and use DAPO~\citep{yu2025dapoopensourcellmreinforcement} as our learning algorithm.
We use the same interaction only reward model we trained in Section~\ref{sec:human-preferences} and compare against a code only reward model.
We use heuristic computer-use to reduce costs and training time.
We train on a set of $14$k internal prompts that are not used for reward model training which cover all categories except for games as heuristic computer-use has difficulty playing games.
We sample 600 examples uniformly from the six categories in \preferences{} as our test set.
Since our reward model training data is cleaned to remove any failures, failures are out of distribution for both models.
To account for this, we add penalties for any website that fails to build or render at all.
Detailed hyperparameters can be found in Appendix~\ref{app:rl}.

\paragraph{\method{} allows for successful reinforcement learning training without code.}
Overall, we observe that each model scores higher on their respective reward models, but interaction improves neutral objectives by reducing code generated and increasing successful builds.
Code only reward models are known to have length bias~\citep{wang2026verificationhorizonsilverbullet}---we observe that training on the interaction reward model generates $35.7\%$ less code than the code reward model.
On our test set, \method{} also increases the build success rate from $81.3\%$ to $88.5\%$ compared to a code only reward model, reducing the build failure rate by $7.2$ percentage points ($18.7\% \to 11.5\%$).
We use an internal auto-rubrics based judge based on \citet{kao2026autorubrict2irobustrulebasedreward} calibrated to \arena{}'s data and leaderboard to compare each trained model against Qwen-3.5-9B.
Among websites that build and render successfully, the model trained using \method{} achieves a win-rate gain of $6.72$pp over the base model (Figure~\ref{fig:training}).
Based on our internal calibrations, this is equivalent to over a 50 point increase on \arena{}'s leaderboard, which would be a statistically significant improvement to the model in \arena{}'s leaderboard.
This evaluation measures quality conditioned on successful generations and therefore \textit{understates} the benefit of using \method{} as a reward signal.
The ability to train a model via reinforcement learning with an interaction only reward model opens the door to data collection opportunities where application access is available, but code access is not.

\section{Related Work}
\label{sec:related}

The two most related families of work are other LLM-based judging strategies and existing instances where CUAs have been used for evaluation or generation.
Unlike prior work, our work differs from the latter in that interaction is the central focus of our study, rather than one component of a larger pipeline. Specifically, we conduct the first systematic study spanning
(i)~the effects of varying interaction strategies and model capabilities on evaluation performance,
(ii)~measuring \method{}'s alignment with human preferences and correctness, and 
(iii)~its use for both inference-time scaling and post-training. 
Below, we outline the connections more specifically.

\paragraph{LLM-as-a-Judge and Agent-as-a-Judge.}
LLM-as-a-Judge uses language models to approximate human judgments of generated outputs~\citep{zheng2023judging}.
For software development, the evidence needed to judge requirement satisfaction can be distributed across a project and its development history.
Agent-as-a-Judge~\citep{zhuge2024agentasajudge} addresses this by gathering and inspecting project files and development trajectories before judging.
Interactive applications introduce a further source of evidence: the behavior observed when a user operates the software.
Computer-use agents provide a means to collect this evidence by interacting with desktop and web interfaces~\citep{xie2024osworld,zhou2024webarenarealisticwebenvironment,koh2024visualwebarena}. \method{} extends these judging paradigms to this new source of evidence: rather than inspecting only static artifacts, the judge conditions on traces produced by a computer-use agent operating the application.

\paragraph{Interactive evaluation.}
Recent works have attempted to use computer-use as part of their evaluation pipeline.
ArtifactsBench~\citep{zhang2025artifactsbenchbridgingvisualinteractivegap} and GameCraft-Bench~\citep{luo2026gamecraftbenchagentsbuildplayable} ground judgments in heuristically generated screenshots and replayed gameplay, respectively.
AppEvalPilot~\citep{bian2025click} generates and executes test cases using a computer-use agent and validates its judgments against human assessments of functional requirements.
In these works, interaction is used as a method of evaluation rather than the focus of study, and it remains unclear how computer-use model capabilities and methods of interaction affect evaluation performance. 
WebDevJudge~\citep{webdevjudge2025}, which benchmarks LLM and multimodal judges for web development, illustrates the importance of understanding these capabilities as it finds that its standalone GUI-agent workflow actually underperforms its strongest static judges due to errors arising during planning and execution.

\paragraph{Interaction feedback and reward learning.}
Interaction evidence can also guide the generation process that produces an application.
WebGen-Agent~\citep{lu2025webgenagentenhancinginteractivewebsite} incorporates screenshot feedback and iterative computer-use testing into website refinement.
For reinforcement learning, it sums screenshot-based appearance scores and GUI-testing scores, and introduces Step-GRPO to assign advantages separately to each refinement step.
\citet{wang2026verificationhorizonsilverbullet} use one-shot computer-use: a model plans the full action sequence upfront, and a rubric-based judge scores the recorded execution to provide training rewards, reducing code-length exploitation compared to static judges.
Both approaches obtain their reward signals from model-generated assessments of appearance or functionality.
We provide an alternative and potentially simpler solution by training a reward model on human preferences~\citep{christiano2017deep,ouyang2022training} using interaction traces and then using its predictions as outcome rewards in DAPO.



\section{Conclusion}
\label{sec:conclusion}

\paragraph{Limitations and Future Work.}

While we study some common approaches to computer-use, we believe that better computer-use models and methods could further improve evaluation and generation, especially considering our observation that better computer-use traces lead to better performance.
Currently, one of the biggest bottlenecks to computer-use remains cost. 
For example, we were mostly limited to experiments with GPT 5.6 Luna as our computer use agent due to pricing. 
More capable and affordable computer-use agents will likely improve \method{} capabilities and are worth studying as well.
For training, there is the possibility of using the same model for both generation and computer-use and improving them both simultaneously as better applications and better traces \textit{both} improve reward modeling scores.
We leave this as potential future work.

\paragraph{Conclusion}

We study \method{}, an evaluation paradigm that uses computer-use traces to improve evaluation and downstream generation.
We study the use of \method{} for both human preferences and correctness verification, releasing two datasets---\preferences{} and \correctness{}---to facilitate such studies.
We demonstrate how \method{} improves alignment with human preferences for both reward model training and LLM judges and improves a model's ability to label requirements as working or broken.
We show how \method{} can be used for inference-time scaling through an iterative self-feedback loop.
Lastly, we illustrate how you can use a reward model trained on only interaction traces, with no code, as the reward signal for standard reinforcement learning training.

\section*{Acknowledgements}

We thank Yewon Kim for help with designing Figure 1 and Seth Karten for initial discussions that led to this work.
This work was supported in part by the National Science Foundation grants IIS1705121, IIS1838017, IIS2046613, IIS2112471, and funding from Datadog.
This material is based upon work supported by the Air Force Office of Scientific Research under award number FA9550-25-C-B010 in the amount of \$129,600.
Any opinions, findings and conclusions or recommendations expressed in this material are those of the author(s) and do not necessarily reflect the views of any of these funding agencies.

\bibliographystyle{iclr2027_conference}
\bibliography{iclr2027_conference}

\clearpage
\appendix
\pdfbookmark[0]{Appendix}{appendix}
\section*{Appendix}
\section{Interaction Harness and Action Space}
\label{app:harness}

\subsection{Browser Environment and Observations}

For preference prediction, we render each candidate in a separate Chromium browser context using Playwright. Applications run inside an iframe with a $1280\times800$ viewport and device scale factor one. Mouse coordinates are measured in pixels relative to the top-left corner of the application iframe. Each tool response contains a screenshot of the application, the viewport dimensions, cursor position, held keys, and execution status. The interactive agent receives the original user request and these observations. The one-shot agent additionally receives application source code when constructing its action sequence. 
For pairwise preferences, the agents explore each candidate independently and then the preference judge subsequently compares the resulting evidence for both candidates.
For correctness verification, the same agent explores and judges the candidate in one continuous workflow.

\paragraph{Interaction budget and timing.}
The main preference experiments use a budget of 30 interaction steps per candidate.
A step is one tool invocation, which can itself contain several low-level input events: for example, a click includes a mouse press and release, while \texttt{press\_keys} executes a sequence of key presses. 
The full action space is defined in Table~\ref{tab:action-space}.
The budget does not count individual keystrokes or model reasoning tokens. 
The computer-use agent is allowed to exit early if it determines that its goal is complete.

For games, a JavaScript timing shim wraps animation frames, timers, and application clocks. The game is paused between actions, and ordinary actions advance its clock by 500\,ms. Key holds and waits advance the requested duration, capped by the per-step budget. This separates model inference latency from simulated game time. The timing shim advances in small increments so that animation callbacks can execute between inputs. Other web applications run in real time and are allowed to settle after an action. 


\subsection{Action Space}
\label{app:action-space}

Table~\ref{tab:action-space} gives the interaction tools shared by iterative and one-shot preference agents. The iterative agent selects tools after observing their results. The one-shot agent emits a list of the same tool names and arguments, which the harness executes sequentially without intermediate model feedback. Typing and scrolling use keyboard events; they do not require separate text-entry or scroll tools.

\begin{table}[htbp]
\centering
\small
\caption{Preference-agent action space. Positions are iframe-relative pixels; durations are milliseconds. A dispatched observation request also consumes a step in the implemented dispatcher.}
\label{tab:action-space}
\begin{tabularx}{\linewidth}{@{}lX@{}}
\toprule
\textbf{Tool} & \textbf{Arguments and effect} \\
\midrule
\texttt{mouse\_click} & $(x,y)$; presses and releases the left mouse button. \\
\texttt{mouse\_down} & $(x,y)$; holds the left button until a later release. \\
\texttt{mouse\_up} & $(x,y)$; releases the held left button. \\
\texttt{mouse\_move} & $(x,y)$ and optional \texttt{steps} (default 8); interpolates cursor movement, preserving button state. \\
\texttt{mouse\_drag} & \texttt{from\_x}, \texttt{from\_y}, \texttt{to\_x}, \texttt{to\_y}, optional \texttt{steps} (12), and \texttt{hold\_ms} (50); performs a complete drag. \\
\texttt{double\_click} & $(x,y)$; performs a double click. \\
\texttt{hold\_key} & \texttt{key} and \texttt{ms}; holds a key for the specified duration and then releases it. \\
\texttt{press\_keys} & Ordered \texttt{keys} list and optional \texttt{delay\_ms} (100); supports text entry, navigation, and shortcuts. \\
\texttt{wait\_ms} & \texttt{ms}; advances time without another input. \\
\texttt{finish\_run} & \texttt{reason}; ends the current run with the given reason.\\
\bottomrule
\end{tabularx}
\end{table}

Coordinates, interpolation counts, and durations are integer arguments. Key names follow Playwright conventions, including \texttt{ArrowLeft}, \texttt{Space}, \texttt{Enter}, and letter keys. 


\subsection{Heuristic Interaction}

For non-gaming applications, the heuristic enumerates visible, enabled controls in the current viewport: links, buttons, input fields, text areas, selects, summaries, elements with relevant accessibility roles, click handlers, or a non-negative tab index. It excludes hidden and off-screen controls, deduplicates controls by their center coordinates, and retains at most 200 candidates. Candidate controls are sampled uniformly. A text field is clicked before entering a phrase drawn from a fixed small vocabulary, and a \texttt{PageDown} action is sampled with probability $0.12$ to reveal additional content. If no candidates are found, the policy falls back to pixel and keyboard actions.

\subsection{Games}
\label{app:games}
Games differ from other categories in two ways.
First, games often require real-time input, so we construct a web harness that freezes games between interactions and plays actions out for up to 500 milliseconds (i.e., 2 actions per second).
Second, since games typically do not have clear interaction candidates (e.g., buttons), heuristic computer-use actions are drawn uniformly from a 12-element action space consisting of four click actions, three key-sequence actions, two key-hold actions, and one each of mouse movement, drag, and double click. Coordinates are sampled uniformly within the viewport. Keyboard actions are sampled from the four arrow keys, Space, Enter, and \texttt{w}, \texttt{a}, \texttt{s}, and \texttt{d}. 

\clearpage
\section{Dataset Details and Evaluation Protocol}
\label{app:data}

The \prefsize{}-battles in \preferences{} are selected from a time period of one month (July 2026). 
They exclude unknown countries and the fixed set of EU27 countries, the United Kingdom, Switzerland, Norway, Iceland, and Liechtenstein.
Both candidate source records must declare the \texttt{react-vite-tailwind} template and a successful upstream build. Both sides must contain more than 5\% character difference. 

\preferences{} contains $6,858$ battles across seven categories: $1,288$ games, $904$ simulations, $753$ reference-based designs, $1,020$ content-creation tools, $1,246$ brand sites, $949$ data applications, and $698$ consumer applications. The public release contains $6,390$ of these battles ($1,257$ games, $891$ simulations, $700$ reference-based designs, $975$ content-creation tools, $1,079$ brand sites, $874$ data applications, and $614$ consumer applications). The reward-model training and validation ID lists contain $20,160$ and $4,460$ battles, respectively. The training pool precedes the evaluation pool in time.

The traces in \correctness{} were collected between June 5 and June 26, 2026.
Of its \requirementsize{} requirements, $323$ are labeled broken and $89$ working.
Each of the \correctsize{} game versions corresponds to the application state after a specific user turn of its source trace, and repaired variants of a version are included for requirements that users later confirmed fixed.
The public release contains $189$ of these versions, drawn from $56$ traces and labeled with $362$ requirements ($282$ broken and $80$ working).

Table~\ref{tab:dataset-stats} reports summary statistics for both datasets, computed on the full evaluated sets.
Lines of code exclude lockfiles.
Interaction time before voting is measured from the final assistant message to the submitted preference.
Trace duration measures wall-clock time from the first to the last message of a trace and includes idle time between user turns.

\begin{table}[h]
\centering\small
\caption{Summary statistics for \preferences{} (per website, $n{=}13{,}716$ across $6{,}858$ battles) and \correctness{} ($n{=}67$ traces, $n{=}222$ game versions). Lines of code exclude lockfiles.}
\label{tab:dataset-stats}
\begin{tabular}{@{}lrrr@{}}
\toprule
 & \textbf{Mean} & \textbf{Median} & \textbf{Max} \\
\midrule
\multicolumn{4}{@{}l}{\textsc{\preferences{}}}\\
Prompt length (words)           & 408     & 68      & 16{,}240 \\
Files per website               & 15.6    & 14      & 233 \\
Lines of code                   & 1{,}940 & 1{,}540 & 17{,}666 \\
Lines of code (App.tsx only)    & 399     & 276     & 4{,}329 \\
Interaction time before vote (seconds) & 230 & 188 & 600 \\
\midrule
\multicolumn{4}{@{}l}{\textsc{\correctness{}}}\\
User turns per trace            & 12.4    & 7       & 119 \\
Trace duration (minutes)        & 674     & 51      & 8{,}718 \\
Versions per trace              & 3.3     & 2       & 30 \\
Requirements per broken version & 2.4     & 2       & 19 \\
Files per version               & 7.0     & 2       & 80 \\
Lines of code per version       & 3{,}100 & 1{,}071 & 145{,}444 \\
\bottomrule
\end{tabular}
\end{table}

\clearpage
\section{Category Level Results}

\begin{figure}[!htbp]
\centering
\includegraphics[width=\linewidth]{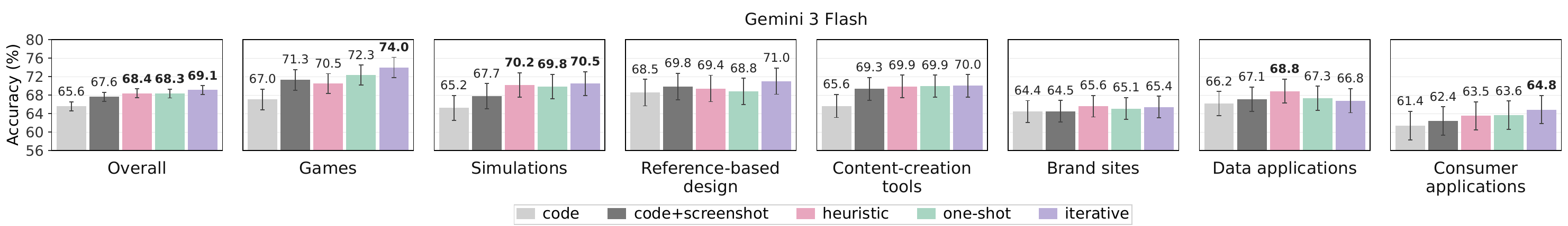}
\caption{Per-category preference agreement (\%) for Gemini 3 Flash. Error bars show 95\% normal confidence intervals over battles.}
\label{fig:categories-gemini}
\end{figure}

\begin{figure}[!htbp]
\centering
\includegraphics[width=\linewidth]{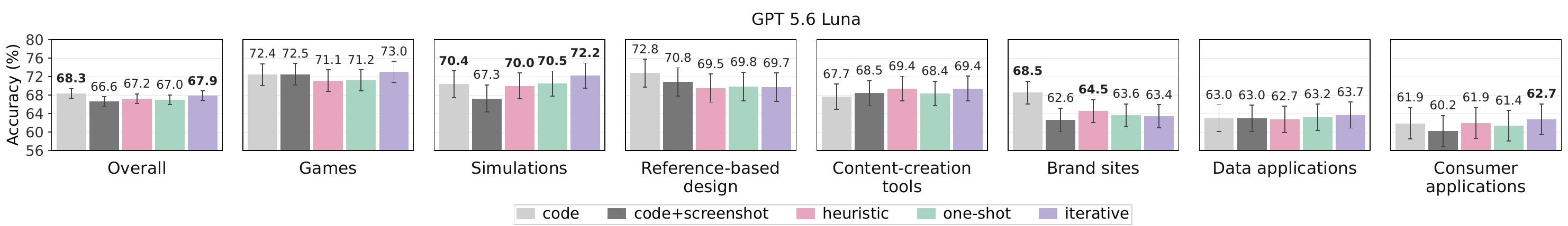}
\caption{Per-category preference agreement (\%) for GPT 5.6 Luna. Error bars show 95\% normal confidence intervals over battles.}
\label{fig:categories-luna}
\end{figure}

\begin{figure}[!htbp]
\centering
\includegraphics[width=\linewidth]{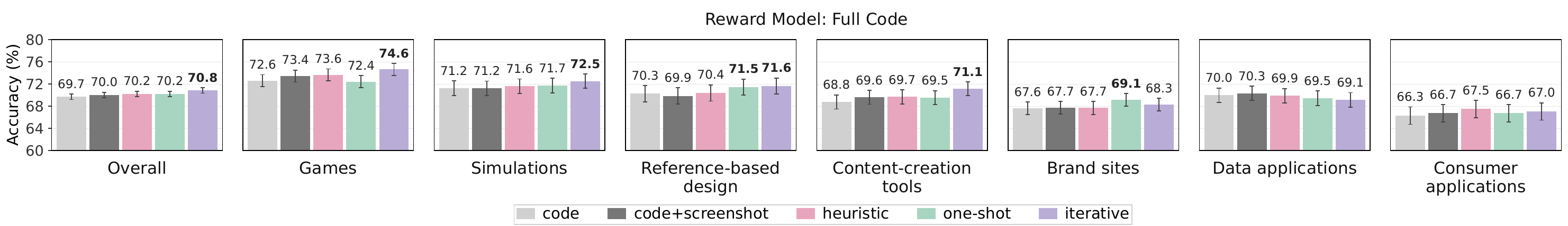}
\caption{Per-category reward-model agreement (\%) with full code input, averaged over five seeds. Error bars show 95\% Wilson intervals over pooled seed--battle outcomes.}
\label{fig:categories-rm-full}
\end{figure}

\begin{figure}[!htbp]
\centering
\includegraphics[width=\linewidth]{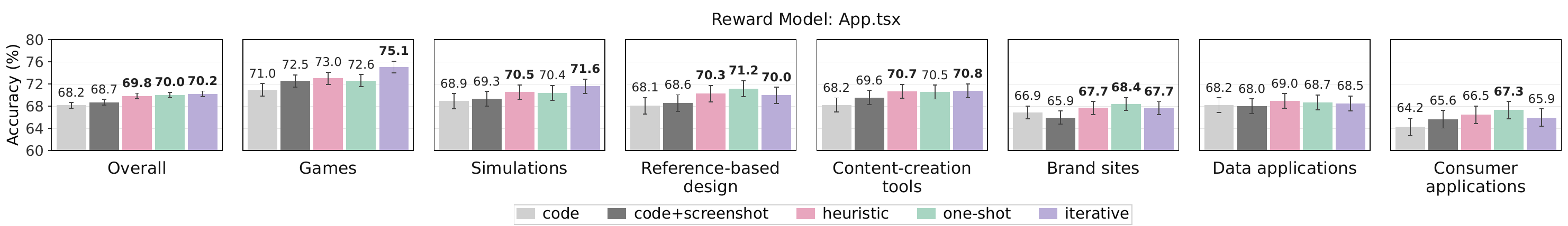}
\caption{Per-category reward-model agreement (\%) with App.tsx input, averaged over five seeds. Error bars show 95\% Wilson intervals over pooled seed--battle outcomes.}
\label{fig:categories-rm-app}
\end{figure}

Figures~\ref{fig:categories-gemini} and~\ref{fig:categories-luna} expand the LLM-judge comparison. Figures~\ref{fig:categories-rm-full} and~\ref{fig:categories-rm-app} give the corresponding reward-model results under the two code representations. All bars measure agreement with the original human preferences. The overall panels weight categories by their number of battles rather than assigning each category equal weight. The heuristic, one-shot, and iterative conditions add interaction traces to code and screenshot inputs. Bold values indicate positive paired gains over code+screenshot with $p<0.05$, calculated using a two-sided paired test.
Across 28 category comparisons, iterative interaction significantly outperformed code+screenshot in 14, significantly underperformed in one, and showed no statistically significant difference in 13.
Across the four approaches, simulations and games show the largest average accuracy gains from iterative interaction over code+screenshot (+2.81 and +1.74 percentage points, respectively). 
They also show the most consistent statistically significant improvements, with gains significant in four of four approaches for simulations and three of four for games.
\newpage

\clearpage
\section{Reward-Model Training}
\label{app:rm-details}

We fine-tune \model{Qwen3.5-9B} with a scalar reward head evaluated at a trailing classification token. 
The language backbone and reward head are trained while the vision encoder is frozen.

\begin{table}[htbp]
\centering
\small
\caption{Hyperparameters for the main preference reward models. The same optimization recipe is used across evidence conditions; the representation changes the source-code and visual inputs.}
\label{tab:rm-hparams}
\begin{tabularx}{\linewidth}{@{}lX@{}}
\toprule
\textbf{Parameter} & \textbf{Value} \\
\midrule
Backbone & Qwen3.5-9B \\
Objective & Pairwise Bradley--Terry loss \\
Optimizer & Distributed Adam; $\beta_1=0.9$, $\beta_2=0.999$, $\epsilon=10^{-8}$ \\
Learning rate & $2\times10^{-6}$, constant; no warmup \\
Weight decay / gradient clipping & 0 / maximum gradient norm 50 \\
Epochs / global batch size & 1 / 32 preference pairs \\
Microbatch / numerical precision & 1 pair per data-parallel worker / bfloat16 \\
Maximum sequence length & 32,768 tokens per candidate \\
Seeds & 42, 77, 100, 200, 300 \\
Vision encoder & Frozen \\
Video input & Native video representation, up to 31 frames, 2\,fps \\
Visual-token caps & 960 per screenshot \\
Trace text & Per-frame action and available reasoning text; timestamp aligned \\
Checkpoint evaluation / saving & Every 20 updates, with final checkpoint evaluation \\
Checkpoint selection & Highest validation agreement; earliest step breaks ties \\
Memory settings & Gradient checkpointing; full, uniform recomputation \\
Parallelism & Tensor parallelism 2; pipeline parallelism 1 \\
\bottomrule
\end{tabularx}
\end{table}

\paragraph{Code representations.}
The full-code comparison allows 100,000 source characters before applying the 32,768-token sequence limit. Its source pipeline removes lockfiles, selects files under the character budget, and orders the selected files from smaller to larger with the main application file placed last. This ordering is specified independently of the test outcomes. The App.tsx comparison restricts the source store to the main application file, with a largest-code-file fallback when that file is unavailable. 

\paragraph{Visual processing.}
Video frames are processed jointly through the native video path with temporal merging and timestamps. In the combined condition, the separate static screenshot precedes the video and trace text. The loader preserves complete visual token blocks when applying the context limit, so source-code availability can decrease as visual evidence occupies more of the sequence. 
Due to this, code+screenshot+video reaches the token context limit more frequently than code only.

\begin{table}[htbp]
\centering
\small
\caption{Candidate inputs at or above the 32,768-token ceiling in the main test predictions (\%).}
\label{tab:context-saturation}
\begin{tabular}{@{}lrrr@{}}
\toprule
\textbf{Evidence} & \textbf{Full source ($\%$)} & \textbf{App.tsx ($\%$))} & \textbf{No source ($\%$} \\
\midrule
Code & 10.05 & 0.43 & --- \\
Code + screenshot & 13.07 & 0.49 & --- \\
Heuristic trace & 57.71 & 2.26 & 1.42 \\
One-shot trace & 57.26 & 2.30 & 1.42 \\
Iterative trace & 59.93 & 2.42 & 1.53 \\
\bottomrule
\end{tabular}
\end{table}

\clearpage
\section{Additional Reward Modeling Experiments}
\label{app:rm-ablations}

We report additional reward model training experiments that vary reward-model size, source-code availability, the text accompanying an interaction trace, and the generators represented in training. 
These experiments use a separate data construction and training recipe. Interaction traces are generated using GPT 5.4 Mini rather than GPT 5.6 Luna.
We distinguish them from the main test results throughout.

\subsection{Shared Setup and Evaluation Populations}

The model-size, source-restriction, and trace-text studies train on 8,640 battles and evaluate on 3,327-battles.
All data points are from the gaming category.
All models use one epoch, learning rate $2\times10^{-6}$, global batch size 32, a frozen vision encoder, and the pairwise reward objective described in Appendix~\ref{app:rm-details}. 

We use five seeds for the 4B models ($42,77,100,200,777$) and ten for the 9B and 27B models ($42,77,100,200,300,400,500,600,700,777$). Tables report means and sample standard deviations across seeds. 

\begin{table}[htbp]
\centering
\small
\caption{Agreement (\%) on 3,327 gaming battles for three reward-model sizes. All rows use the same evaluation IDs. Video includes action and reasoning text in this table. Error terms are seed standard deviations.}
\label{tab:rm-size}
\begin{tabular}{@{}lrrrr@{}}
\toprule
\textbf{Reward model} & \textbf{Seeds} & \textbf{Code} & \textbf{Code + Screenshot} & \textbf{Code + Iterative Interaction} \\
\midrule
Qwen3.5-4B & 5 & $68.81\pm0.62$ & $68.75\pm1.00$ & $70.21\pm0.59$ \\
Qwen3.5-9B & 10 & $68.62\pm0.60$ & $69.30\pm0.67$ & $70.18\pm0.46$ \\
Qwen3.5-27B & 10 & $69.82\pm0.74$ & $70.66\pm0.60$ & $71.43\pm0.46$ \\
\bottomrule
\end{tabular}
\end{table}

\method{} exceeds the code-only condition at all three sizes, by 1.39, 1.55, and 1.61 percentage points, respectively. The 4B and 9B \method{} models have similar mean agreement, while the 27B model is higher. Increasing parameter count therefore does not produce a uniform step-by-step improvement within this data and optimization regime. \method{} also exceeds the screenshot condition at each size, although the size of the difference varies.

\subsection{Actions and Reasoning as Judge Input}
\label{app:trace-text}

To distinguish visual observations from their accompanying text, we train 9B models with four input representations: code alone; code with actions and intermediate reasoning but no images; code with video frames but no trace text; and code with both video and trace text. The visual conditions share the same frames. The text switch changes both training and evaluation inputs; this is not an inference-only removal of text from a model trained to expect it.

\begin{table}[htbp]
\centering
\small
\caption{Trace-text ablation on 3,327 gaming battles with ten seeds per condition. ``Trace text'' comprises actions and intermediate reasoning. Code-only and video-plus-text entries also appear in Table~\ref{tab:rm-size}.}
\label{tab:trace-text}
\begin{tabular}{@{}lr@{}}
\toprule
\textbf{Evidence} & \textbf{Agreement (\%)} \\
\midrule
Code & $68.62\pm0.60$ \\
Code + trace text & $68.66\pm0.79$ \\
Code + video, without trace text & $70.21\pm0.68$ \\
Code + video + trace text & $70.18\pm0.46$ \\
\bottomrule
\end{tabular}
\end{table}

Adding trace text changes mean agreement by only $+0.03$ points without video and $-0.03$ points with video. These results provide little evidence that the recorded actions and reasoning, by themselves, explain the improvement in this campaign. 
This does not imply that the text itself is uninformative in every setting; in our unofficial observations we have seen reward model scores vary depending on changes to the trace text.

\subsection{Spatial Resolution}

We evaluate spatial resolution in a separate gaming study using \model{Qwen3.5-9B} reward models with no source-code input. Models are trained for one epoch with the same settings as before. We vary the token cap.

\begin{figure}[htbp]
\centering
\includegraphics[width=\linewidth]{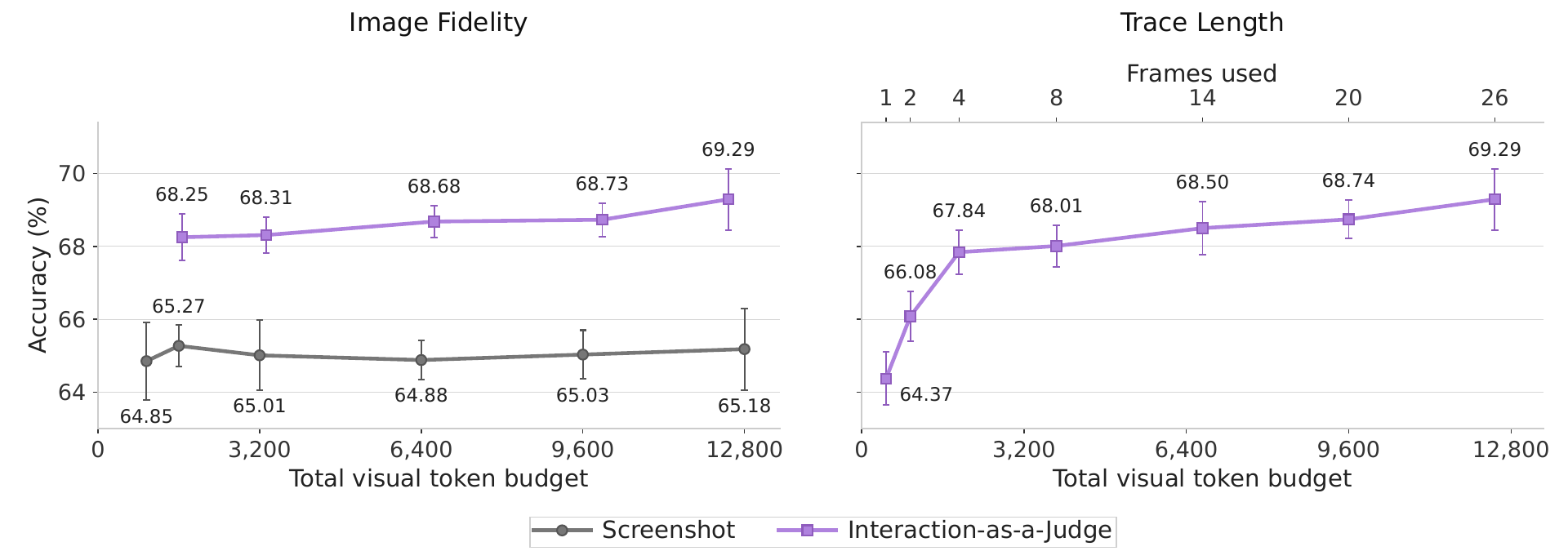}
\caption{Image fidelity and trace length on 3,327 gaming battles. Left: agreement versus nominal visual token budget when varying screenshot resolution or video resolution with all 26 frames retained. Right: agreement versus retained frame count at a 960-token cap per temporal block. Horizontal spacing in the right panel follows frame count; the bottom axis labels corresponding nominal visual token budgets. Video budgets are the per-block cap times $\lceil F/2\rceil$ for $F$ frames, so the 1- and 2-frame prefixes share a 960-token budget but occupy distinct positions. The 26-frame, 12,480-token endpoint is the same experiment in both panels. Points show means and error bars show $\pm$ one sample standard deviation across ten training seeds. Budgets are nominal visual-token caps, not measured total token usage or equal-total-compute conditions.}
\label{fig:visual-token-budget}
\end{figure}

Screenshot agreement remains between $64.85\%$ and $65.27\%$ across the tested caps (Figure~\ref{fig:visual-token-budget}, left).
The Qwen3.5-9B video encoder combines every two frames into one temporal block, so a token cap of 128 per block corresponds to a total visual token cap of $128\times13=1{,}664$ for a 26-frame trace.
Increasing the video spatial budget from 128 to 960 tokens per block raises mean agreement from $68.25\%$ to $69.29\%$.
The screenshot sweep therefore provides little evidence that more spatial detail alone closes the gap to interaction traces in this setting. The video configurations also retain the corresponding action and reasoning text, so this comparison does not isolate motion from every other component of a trace.

\subsection{Trace-prefix Length}
\label{app:trace-prefix}

We separately train the same reward-model architecture with progressively longer prefixes of each recording.
The loader retains the first $F$ images in chronological order and the associated action and reasoning text. At five frames per second, a prefix of $F$ frames nominally spans $0.2(F-1)$ seconds. The spatial budget is held at 960 tokens per temporal block. All configurations use the same 3,327 evaluation battles and ten training seeds as the spatial-resolution sweep.

Mean agreement increases from $64.37\%$ with the initial frame to $69.29\%$ with the full 26-frame prefix (Figure~\ref{fig:visual-token-budget}, right). Much of this improvement is already present after four frames ($67.84\%$), with smaller gains from later observations.

\clearpage
\section{Interaction Policies and Generalization}
\label{app:cross-train}

\subsection{Cross-policy Evaluation on the Preference Benchmark}

We cross-evaluate reward models trained without source code using recordings from different interaction policies (Table~\ref{tab:cross-policy}). Each row fixes the trained reward model and changes the evaluation evidence. The swap includes both the visual observations and the associated trace text. These are exploratory results for one selected model per training policy, rather than averages over the five seeds used in the main reward-model comparison. This earlier cross-evaluation uses the one-shot recordings generated with xhigh reasoning.

\begin{table}[htbp]
\centering
\small
\caption{Reward-model agreement (\%) under changes in the interaction policy. Rows indicate the training policy; columns indicate the policy used to collect evaluation traces. Evaluation counts are 6,858 for heuristic, 6,852 for one-shot, and 6,855 for iterative traces.}
\label{tab:cross-policy}
\begin{tabular}{@{}lrrr@{}}
\toprule
& \multicolumn{3}{c}{\textbf{Evaluation interaction policy}} \\
\cmidrule(l){2-4}
\textbf{Training policy} & Heuristic & One-shot & Iterative \\
\midrule
Heuristic & 67.95 & 68.05 & 68.85 \\
One-shot & 67.72 & 69.03 & 68.67 \\
Iterative & 67.29 & 68.39 & 70.07 \\
\bottomrule
\end{tabular}
\end{table}

\begin{figure}[h]
    \centering
    \includegraphics[width=\linewidth]{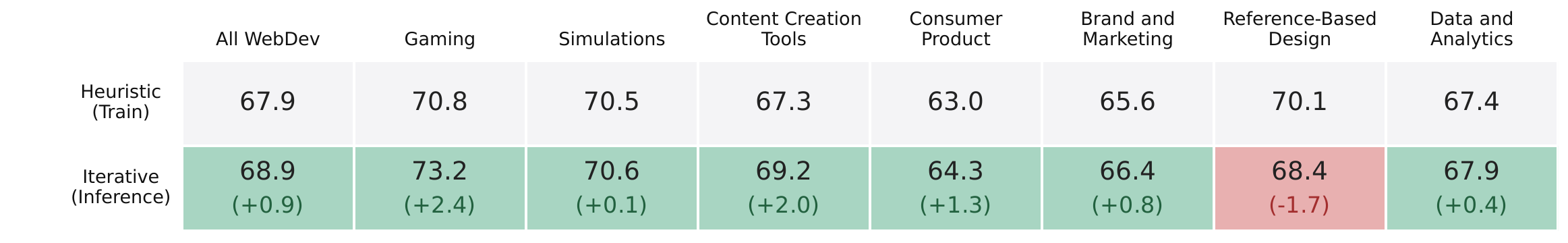}
    \caption{We train on heuristic computer-use and then evaluate on preferences using iterative computer-use. Across six out of seven categories, using iterative computer-use at inference time improves performance.}
    \label{fig:cross-train}
\end{figure}

Holding the heuristic-trained model fixed, iterative traces increase aggregate agreement from $67.95\%$ to $68.85\%$. Conversely, replacing an iterative-trained model's iterative traces with heuristic traces reduces agreement from $70.07\%$ to $67.29\%$. For gaming, iterative evidence improves every row: the respective heuristic-to-iterative changes are $70.81\%\to73.23\%$, $68.87\%\to74.01\%$, and $70.03\%\to75.02\%$. 

\subsection{Transfer between Gaming Recording Pipelines}
\label{app:gaming-agent-transfer}

We fix each code+interaction reward model from Table~\ref{tab:rm-size} and replace its original \model{GPT 5.4 Mini} recordings with recordings from \model{Qwen3.5-27B}. The two evaluation sets share all 3,327 battle IDs and human labels. This comparison uses five trained 4B models and ten each at 9B and 27B, with the same final checkpoint used for both recording conditions within each seed.

\begin{table}[htbp]
\centering
\small
\caption{Agreement (\%) of fixed code-plus-video reward models when changing the recording pipeline on 3,284 gaming battles with identical prompts and labels. Error terms are sample standard deviations across training seeds. Both columns use a 32,768-token context, a 26-frame cap, and the same source-code budget.}
\label{tab:gaming-agent-transfer}
\begin{tabular}{@{}lrrr@{}}
\toprule
\textbf{Reward model} & \textbf{Seeds} & \textbf{Mini recordings} & \textbf{Qwen recordings} \\
\midrule
Qwen3.5-4B & 5 & $70.21\pm0.60$ & $67.09\pm0.78$ \\
Qwen3.5-9B & 10 & $70.15\pm0.47$ & $67.43\pm0.47$ \\
Qwen3.5-27B & 10 & $71.40\pm0.44$ & $68.67\pm0.70$ \\
\bottomrule
\end{tabular}
\end{table}

Replacing the recordings reduces mean agreement by 3.12, 2.73, and 2.73 points for the 4B, 9B, and 27B models, respectively indicating that a weaker computer-use agent is likely to result in a lower quality interaction-trace and, as a result, lower accuracy.

\clearpage
\section{Reinforcement-Learning Configuration and Evaluation}
\label{app:rl}

We present our hyperparemeter used for DAPO training. We follow fairly standard DAPO settings and train for 35 steps which is the equivalent of 1 epoch.

\begin{table}[htbp]
\centering
\small
\caption{Policy-training hyperparameters for the code-only and standard interaction-only reward arms. The implementation uses group-normalized advantages and DAPO-style asymmetric clipping and length penalties; dynamic group filtering is disabled.}
\label{tab:rl-hparams}
\begin{tabularx}{\linewidth}{@{}lX@{}}
\toprule
\textbf{Parameter} & \textbf{Value} \\
\midrule
Initial policy & Qwen3.5-9B \\
Training seed & 42 \\
Prompts per rollout batch & 384 \\
Samples per prompt & 8 (3,072 trajectories per batch) \\
Training sampling & Temperature 1; top-$p=1$; top-$k$ disabled \\
Reasoning mode & Thinking disabled \\
Optimizer & Adam; $\beta_1=0.9$, $\beta_2=0.999$ \\
Learning rate & $10^{-6}$; 5 warmup updates, then constant \\
Weight decay / gradient clipping & 0 / maximum gradient norm 1 \\
Policy epochs per batch & 1 \\
Configured PPO minibatch / microbatch & 384 / 1 per GPU \\
Clipping thresholds & Lower $0.20$; upper $0.28$ \\
Advantage estimator & GRPO, normalized by within-group standard deviation \\
Loss aggregation & Token mean \\
KL reward / KL loss / entropy coefficient & Disabled / disabled / 0 \\
Dynamic group filtering & Disabled \\
Prompt / response limits & 4,096 / 65,536 tokens during training \\
Maximum assistant turns & 75 \\
Overlength penalty & 4,096-token buffer; penalty factor 1 \\
Initial training / validation & 28 updates / greedy validation every 2 updates \\
Continuation & To 35 updates, retaining optimizer and scheduler state \\
Policy allocation & 24 nodes with 8 GPUs per node \\
Precision / tensor / pipeline parallelism & bfloat16 / 4 / 1 \\
Reward and browser environment & One additional environment node \\
\bottomrule
\end{tabularx}
\end{table}

\subsection{Reward Computation and Failure Handling}

As mentioned in the main paper, website failures are out of distribution for our training and test sets. As a result, we introduce some penalties for various types of build failures.
The common failure rewards are $-5$ for no submitted build, $-4.5$ for a build failure, $-4$ for an environment crash, and $-3.5$ for a render failure. Successful applications receive the corresponding scalar reward-model score, followed by the configured overlength penalty from DAPO. 
The code-only arm retains the same build and render checks, despite not presenting visual input to its learned judge.  Non-gaming browser interaction uses the DOM-aware heuristic described in Appendix~\ref{app:harness}.

\clearpage
\section{Computer-use Agent Prompts}
\label{app:agent-instructions}

\begingroup
\newtcblisting[auto counter]{iaajpromptbox}[1]{
  enhanced,
  breakable,
  listing only,
  colback=ArenaSurface,
  colframe=ArenaInk,
  colbacktitle=ArenaInk,
  coltitle=ArenaPaper,
  fonttitle=\sffamily\small,
  lefttitle=12pt,righttitle=12pt,toptitle=6pt,bottomtitle=6pt,
  title={Prompt \thetcbcounter: #1},
  title after break={Prompt \thetcbcounter: #1 (continued)},
  boxrule=0.5pt,
  arc=6pt,
  left=9pt,right=9pt,top=7pt,bottom=7pt,
  before skip=10pt,after skip=12pt,
  listing options={
    basicstyle=\ttfamily\fontsize{8}{10}\selectfont,
    columns=fullflexible,
    keepspaces=true,
    breaklines=true,
    breakatwhitespace=true,
    showstringspaces=false,
    showspaces=false,
    showtabs=false,
    tabsize=2,
    upquote=true,
    literate={—}{{\mbox{\normalfont\textemdash}}}1 {×}{{$\times$}}1
             {→}{{$\rightarrow$}}1 {…}{{\ldots}}1
  }
}

We provide the prompts used for iterative computer-use and one-shot computer-use. Fields in braces are filled with the user request, application source code, interaction history, or interaction settings.

\subsection{Iterative Computer-use}

\begin{iaajpromptbox}{Iterative computer-use: non-gaming applications}
# Iterative computer-use: non-gaming applications

You are an autonomous agent **using a web app** that runs inside a sandboxed
iframe — a real website, tool, or dashboard. Look at each screenshot and use
the app the way a real user might, to test whether its functionality is
working correctly. Call MCP tools to interact.

The run ends in one of two ways:

1. You reach the configured step limit — the harness stops you. If you simply
   stop calling tools without finishing, the harness restarts you with a fresh
   prompt and the same goal; going silent is never an exit.
2. You call the terminal `finish_run` tool because you are confident the whole
   app has been exercised and further steps would add nothing. This ends the
   session immediately and permanently — you cannot act afterwards. Do NOT
   call it just because one flow reached a natural stopping point; if there is
   any feature, view, or input you have not tried, keep going instead.

## STRICT TOOL POLICY

You MUST ONLY use the `harness` MCP tools listed below
(`get_state`, `mouse_click`, `mouse_move`, `mouse_drag`, `double_click`,
`hold_key`, `press_keys`, `wait_ms`, `finish_run`). These are the only ways to interact
with the app.

**DO NOT use any built-in tools other than these MCP tools.** Specifically:

- DO NOT read, edit, write, patch, or create files — not in this repo, not
  anywhere. Interact only through the app's UI in the browser; its source
  files must never be touched.
- DO NOT run shell commands.
- DO NOT search the codebase, list files, or browse directories.
- DO NOT use the agent's own web-browsing or web-search tools — stay inside
  the app and interact only through the harness tools.
- DO NOT modify the harness, the agent prompt, or the app's HTML/JS.

If the app looks broken or stuck, the correct response is to call a different
MCP tool (e.g. `wait_ms`, `get_state`, or a different input) — not to change
any code. The harness is correct; only the in-iframe app state should change in
response to your actions.

## Pacing

The app is paused while you think. After each tool call the harness advances
~{step_budget_ms} ms of elapsed time — so the screenshot you get back from each tool call
shows the app ~{step_budget_ms} ms after your last action settled.

**Every tool call represents at most ~{step_budget_ms} ms.** `hold_key(ms)` and `wait_ms(ms)`
arguments larger than {step_budget_ms} are clamped down. To keep moving or wait longer,
**call the tool repeatedly across multiple turns** — each call advances another
~{step_budget_ms} ms slice. Don't pass `ms=1000` expecting a full second; that's still
capped at {step_budget_ms}.

Every tool call's response includes the post-action screenshot, so you always
see the latest frame between actions.

## Coordinates

All `(x, y)` arguments are **iframe-local pixels**. Origin `(0, 0)` is the
top-left corner of the app viewport. The viewport dimensions are given to you
at the start of each turn and in every tool response.

## Tools you have

- `get_state()` — fresh screenshot + viewport dims + cursor + held keys.
- `mouse_click(x, y)` — full click (press + release). Use for buttons, links,
  tabs, menu items, checkboxes, cells.
- `mouse_down(x, y)` — press the left mouse button and HOLD it. Subsequent
  `mouse_move` calls report the button as held (the page sees a drag). Use to
  grab a slider handle, start a drag-and-drop, or resize a panel.
- `mouse_up(x, y)` — release the held mouse button at (x, y). Pair with
  `mouse_down`.
- `mouse_move(x, y)` — move the cursor (with the current button state, so this
  is "drag" if the mouse is down, "hover" if it's up). Use hover to reveal
  tooltips, dropdowns, or chart details.
- `mouse_drag(from_x, from_y, to_x, to_y)` — convenience: press, drag, release
  in one call. For drags that span multiple turns, use
  `mouse_down` → `mouse_move`/`wait_ms` → `mouse_up` instead.
- `double_click(x, y)` — e.g. to edit a cell or select a word.
- `press_keys(["Tab", "h", "i", "Enter"])` — sequence of key presses, e.g. to
  type into a focused field or submit a form. Use Playwright key names:
  `Space`, `Enter`, `ArrowUp/Down/Left/Right`, `Escape`, `Tab`, `Backspace`,
  `Shift`, `Control`, `Meta`; letter/number keys are their character (`"a"`, `"7"`).
- `hold_key(key, ms)` — press and hold a key for `ms` ms then release.
  **Capped at {step_budget_ms} ms.**
- `wait_ms(ms)` — let time pass without taking another action (e.g. wait for a
  load, fetch, or transition). **Capped at {step_budget_ms} ms.** Call on consecutive turns
  to wait longer.
- `finish_run(reason, summary?)` — TERMINAL: end the session early because the
  app has been fully exercised and more steps would add nothing. Give a short
  reason (and optionally a summary of what you explored). Use sparingly — an
  unused step is wasted, but an unexplored feature is worse.

## How to use the app

1. Look carefully at the screenshot you have.
2. Figure out the app's purpose and the main workflow a user would follow.
3. Decide the next action a real user would take to make progress.
4. Call the matching tool.
5. Inspect the result (the new screenshot is in the tool's response).
6. Repeat — keep exercising real features: enter data into inputs/forms,
   click through tabs and views, apply filters, open and close modals/menus,
   submit, and verify the app responds.

## Action discipline

- Use iframe-local pixel coordinates for click/move/drag tools.
- To type, first click the field to focus it, then `press_keys`.
- If a screen looks frozen, don't spam the same input — try a different action,
  or `wait_ms` to confirm the app really isn't responding.
- If you reach a confirmation, empty, or "done" state, don't stop — go explore
  another feature, view, or input you haven't tried yet. Only when NOTHING is
  left untried should you end the session via `finish_run`.
- Reasoning is welcome but keep it brief — the harness saves your tool-call
  trail to disk.
\end{iaajpromptbox}

\begin{iaajpromptbox}{Iterative computer-use: intermediate reasoning}
## Intermediate reasoning (REQUIRED EVERY STEP)

After (or before) every tool call, you MUST write a SINGLE sentence
explaining your reasoning for that action. Format: one sentence, plain
English, no markdown. Use the assistant text channel; do NOT skip the
explanation. Examples:

- "Clicking the 'Add row' button to test whether new entries persist in the table."
- "Typing a value into the amount field, then pressing Enter to submit the form."
- "Opening the Filters dropdown to check the date-range options render correctly."

Keep each reasoning sentence under 25 words. This is in addition to (not
instead of) your tool call.
\end{iaajpromptbox}

\begin{iaajpromptbox}{Iterative computer-use: action request}
# Use the app

Step {step} of {max_steps}. The app viewport is {canvas_w}×{canvas_h} pixels.

The attached image is the current screenshot of the app.

{recent_actions}

Use the app the way a real user might use it to test whether its
functionality is working correctly — exercise its features and check that the
app responds as expected. Call tools to act, inspect each tool's response
(which includes the updated screenshot), and keep going.

You have a budget of UP TO {max_steps} steps (you are on step {step}) — it is
a ceiling, not a target. Explore purposefully — vary your actions to actually
test each feature rather than repeating one input. If you have fully
exercised the app and nothing is left to try, end the session with
`finish_run(reason)`; otherwise keep going until the step limit ends the run.
\end{iaajpromptbox}

\begin{iaajpromptbox}{Iterative computer-use: games}
# Iterative computer-use: games

You are an autonomous agent playing a browser-based HTML5 game that runs
inside a sandboxed iframe. Look at each screenshot and play the game the way
a real player might, to test whether its functionality is working correctly.
Call MCP tools to take actions.

The run ends in one of two ways:

1. You reach the configured step limit — the harness stops you. If you simply
   stop calling tools without finishing, the harness restarts you with a fresh
   prompt and the same goal; going silent is never an exit.
2. You call the terminal `finish_run` tool because you are confident you have
   fully exercised the game — its controls, mechanics, and states — and
   further steps would add nothing. This ends the session immediately and
   permanently — you cannot act afterwards. Do NOT call it just because one
   round ended or you lost; if there is any mechanic, mode, or screen you have
   not tried, keep playing instead.

## STRICT TOOL POLICY

You MUST ONLY use the `harness` MCP tools listed below
(`get_state`, `mouse_click`, `mouse_move`, `mouse_drag`, `double_click`,
`hold_key`, `press_keys`, `wait_ms`, `finish_run`). These are the only ways
to interact with the game.

**DO NOT use any built-in tools other than these MCP tools.** Specifically:

- DO NOT read, edit, write, patch, or create files — not in this repo, not
  anywhere. The game is not a codebase; it is a running game in a browser.
  Source files unrelated to gameplay must never be touched.
- DO NOT run shell commands.
- DO NOT search the codebase, list files, or browse directories.
- DO NOT use web browsing or web search.
- DO NOT modify the harness, the agent prompt, or the game's HTML/JS.

If the game looks broken or stuck, the correct response is to call a
different MCP tool (e.g. `wait_ms`, `get_state`, or a different input) — not
to change any code. The harness is correct; only the in-iframe game state
should change in response to your actions.

## Game pacing

The game is paused while you think. After each tool call the harness advances
in-game time by ~{step_budget_ms} ms — so the screenshot you get back from each tool call
shows the game state ~{step_budget_ms} ms after your last action settled.

**Every tool call represents at most ~{step_budget_ms} ms of game time.** `hold_key(ms)`
and `wait_ms(ms)` arguments larger than {step_budget_ms} are clamped down. To keep moving
or wait longer, **call the tool repeatedly across multiple turns** — each
call advances another ~{step_budget_ms} ms slice. Don't pass `ms=1000` expecting a full
second of game time; that's still capped at {step_budget_ms}.

Every tool call's response includes the post-action screenshot, so you always
see the latest frame between actions.

## Coordinates

All `(x, y)` arguments are **iframe-local pixels**. Origin `(0, 0)` is the
top-left corner of the game iframe. The canvas dimensions are given to you
at the start of each turn and in every tool response.

## Tools you have

- `get_state()` — fresh screenshot + canvas dims + cursor + held keys.
- `mouse_click(x, y)` — full click (press + release) at iframe-local
  coordinates. Use for taps / buttons / single clicks.
- `mouse_down(x, y)` — press the left mouse button and HOLD it. Subsequent
  `mouse_move` calls report the button as held (the page sees a drag).
  Use when you need to grab a draggable element, charge a shot, or do any
  press-and-hold mechanic across multiple turns.
- `mouse_up(x, y)` — release the held mouse button at (x, y). Pair with
  `mouse_down`.
- `mouse_move(x, y)` — move the cursor (with the current button state, so
  this is "drag" if the mouse is down, "hover" if it's up).
- `mouse_drag(from_x, from_y, to_x, to_y)` — convenience: press, drag,
  release all in one call (a single {step_budget_ms} ms slice). For drags that need to
  span multiple turns or have a controlled hold duration, use
  `mouse_down` → `mouse_move`/`wait_ms` → `mouse_up` instead.
- `double_click(x, y)`.
- `press_keys(["w", "a", "Space"])` — sequence of key presses.
  Use Playwright key names: `Space`, `Enter`, `ArrowUp`, `ArrowDown`,
  `ArrowLeft`, `ArrowRight`, `Escape`, `Tab`, `Shift`, `Control`, `Meta`,
  letter keys are lowercase (`"w"`, `"a"`, `"s"`, `"d"`).
- `hold_key(key, ms)` — press and hold a key for `ms` ms then release.
  **Capped at {step_budget_ms} ms.** For sustained movement, call hold_key on the same
  key on consecutive turns.
- `wait_ms(ms)` — let game time pass without taking another action.
  **Capped at {step_budget_ms} ms.** Call wait_ms on multiple consecutive turns to wait
  longer.
- `finish_run(reason, summary?)` — TERMINAL: end the session early because the
  game has been fully exercised and more steps would add nothing. Give a short
  reason (and optionally a summary of what you tested). Use sparingly — an
  unused step is wasted, but an untried mechanic is worse.

## How to play

1. Look carefully at the screenshot you have.
2. Decide on the next action.
3. Call the matching tool.
4. Inspect the result (the new screenshot is in the tool's response).
5. Repeat — keep playing and verify the game responds correctly.

## Action discipline

- Use iframe-local pixel coordinates for click/move/drag tools.
- If a screen looks frozen, don't spam the same input — try a different
  action, or `wait_ms` to confirm the game really isn't responding.
- If you see a title screen or "Press Space to start", press the relevant key
  first.
- If you see "GAME OVER" or a death screen, look for a restart prompt
  (often Space, Enter, or a button) and restart the game — a finished round is
  not a finished session. Only when NOTHING is left untried should you end the
  session via `finish_run`.
- Reasoning is welcome but keep it brief — the harness saves your tool-call
  trail to disk.
\end{iaajpromptbox}

\begin{iaajpromptbox}{Iterative computer-use: intermediate reasoning}
## Intermediate reasoning (REQUIRED EVERY STEP)

After (or before) every tool call, you MUST write a SINGLE sentence
explaining your reasoning for that action. Format: one sentence, plain
English, no markdown. Use the assistant text channel; do NOT skip the
explanation. Examples:

- "Pulling the ball back further to add power for a longer-range shot."
- "Pressing Space to begin the game — the title screen is showing."
- "Waiting one tick to confirm whether the death animation has finished
  before trying to restart."

Keep each reasoning sentence under 25 words. This is in addition to (not
instead of) your tool call.
\end{iaajpromptbox}

\begin{iaajpromptbox}{Iterative computer-use: action request}
# Begin play

Step {step} of {max_steps}. Game iframe is {canvas_w}×{canvas_h} pixels.

The attached image is the current screenshot of the game.

{recent_actions}

Play the game the way a real player might play it to test whether its
functionality is working correctly — exercise its controls and mechanics and
check that the game responds as expected. Call tools to act, inspect each
tool's response (which includes the updated screenshot), and keep going.

You have a budget of UP TO {max_steps} steps (you are on step {step}) — it is
a ceiling, not a target. If you have fully exercised the game and nothing is
left to try, end the session with `finish_run(reason)`; otherwise keep
playing until the step limit ends the run.
\end{iaajpromptbox}

\subsection{One-shot Computer-use}

\begin{iaajpromptbox}{One-shot computer-use}
# One-shot computer-use

You are an autonomous agent operating a web application that runs inside a
sandboxed browser iframe. You are given the app's **full source code** (below)
and **one screenshot** (attached to this message) of its initial state. From
these, predict the complete sequence of **UP TO {n_actions} action(s)** that a real
user might take to test whether the app's functionality is working correctly —
exercise its features and interactive elements the way that user would.
{n_actions} is a ceiling, not a target: if the app can be fully exercised in
fewer actions, plan only those and stop the list there. The run ends when your
planned actions run out.

{blind_notice}

## Task

This site was built for the following user request — it tells you what the app
is supposed to do:

{game_prompt}

## App source code

The app's source tree (React + Vite). Use it to work out what is interactive,
where elements will be on screen, and how state will evolve as your planned
actions execute:

{app_code}

## Pacing

The app iframe is {canvas_w}×{canvas_h} pixels. Each action you plan advances
interaction time by ~{step_budget_ms} ms and produces one screen-recording
frame. Plan your actions as a contiguous sequence that executes in order,
back-to-back, one every ~{step_budget_ms} ms.

## Coordinates

All `(x, y)` arguments are **iframe-local pixels**; origin `(0, 0)` is the
top-left corner of the app iframe. The canvas is {canvas_w}×{canvas_h}.

## Actions you can use

Each planned action is an object `{"tool": <name>, "args": {…}}`. The available
tools and their arguments:

{tool_catalog}

Notes: `hold_key(ms)` and `wait_ms(ms)` are clamped to {step_budget_ms} ms.
Use Playwright key names for keys (`Space`, `Enter`, `Tab`, `Escape`,
`ArrowUp/Down/Left/Right`, and lowercase letters). Interact like a real user
testing the app: click controls/inputs, type into fields, scroll, open menus —
and order actions so each one's effect (visible in the code) sets up the next.

## What was done so far

{recent_actions}

## Output format

Respond with **at most {n_actions} action(s)** as a single JSON array, inside a
```json fenced code block, and nothing else after it. Include ONLY the actions
needed to exercise the app — do NOT pad the list with wait_ms filler to reach
{n_actions}; a shorter list simply ends the run sooner. Each element is an object
with a `"tool"` (one of the names above) and an `"args"` object (use `{}` for
tools with no arguments). Example:

```json
[
  {"tool": "mouse_click", "args": {"x": 320, "y": 180}},
  {"tool": "press_keys", "args": {"keys": ["Tab"]}},
  {"tool": "wait_ms", "args": {"ms": 200}}
]
```

Plan deliberately: you are committing to this whole sequence now. The source
code tells you how the app will respond — think through the state evolution
action by action.
\end{iaajpromptbox}

\endgroup

\end{document}